\documentclass[10pt,twocolumn,letterpaper]{article}

\usepackage{cvpr}
\usepackage{times}
\usepackage{epsfig}
\usepackage{graphicx}
\usepackage{amsmath}
\usepackage{amssymb}

\usepackage{xcolor}
\usepackage{ifthen}
\usepackage{caption}
\usepackage{makecell}
\usepackage{textcomp}
\usepackage{array}
\usepackage{mathtools}
\usepackage{subcaption}

\DeclareMathOperator{\atantwo}{atan2}
\DeclareMathAlphabet\mathbfcal{OMS}{cmsy}{b}{n}

\newcommand{\threesixty}{\text{360$^\circ$}\xspace}

\makeatletter
\renewcommand{\paragraph}{%
  \@startsection{paragraph}{4}%
  {\z@}{0\baselineskip \@plus 0ex \@minus 0ex}{-0em}%
  {\normalfont\normalsize\bfseries}%
}
\makeatother

\definecolor{red}{rgb}{0.9,0.1,0}
\definecolor{slateblue}{rgb}{0.7,0.35,0.9}
\definecolor{green}{rgb}{0, 0.4, 0}
\definecolor{orange}{rgb}{0.5, 0.25, 0}
\definecolor{mahogany}{rgb}{0.75, 0.25, 0.0}
\definecolor{purple}{rgb}{0.3, 0, 0.3}
\definecolor{darkgreen}{rgb}{0, 0.4, 0}
\definecolor{frenchblue}{rgb}{0.0, 0.45, 0.73}
\definecolor{blue}{rgb}{0.0, 0.0, 1.0}
\definecolor{goldenrod}{rgb}{0.85, 0.65, 0.13}
\newboolean{revising}
\setboolean{revising}{false}
\ifthenelse{\boolean{revising}}
{
    \newcommand{\ignore}[1]{}
    \newcommand{\revise}[1]{\textcolor{red}{#1}}

} {
    \newcommand{\ignore}[1]{}
    \newcommand{\revise}[1]{#1}

}

\usepackage[breaklinks=true,bookmarks=false]{hyperref}

\cvprfinalcopy 

\def\cvprPaperID{1073} 

\begin{document}

\newcommand{\modelname}{\text{HorizonNet}\xspace}
\title{\modelname: Learning Room Layout with 1D Representation and Pano Stretch Data Augmentation}

\author{
Cheng Sun \qquad Chi-Wei Hsiao \qquad Min Sun \qquad Hwann-Tzong Chen \\
National Tsing Hua University \\
{\tt\small \{chengsun, chiweihsiao\}@gapp.nthu.edu.tw} \qquad {\tt\small sunmin@ee.nthu.edu.tw} \qquad {\tt\small htchen@cs.nthu.edu.tw}
}

\twocolumn[{%
\maketitle
\renewcommand\twocolumn[1][]{#1}%
   \vspace{-5mm} 
    \centering
    \includegraphics[width=.96\linewidth]{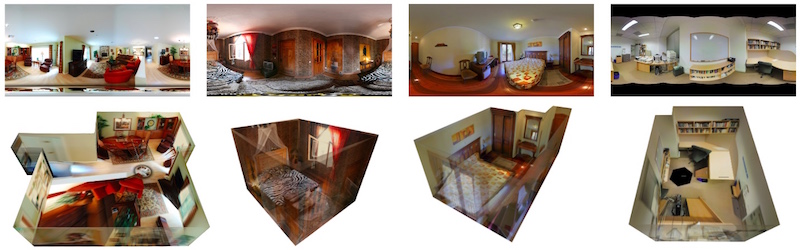}
    \captionof{figure}{Some examples of 3D reconstructed room layouts by our \modelname.\label{fig:teaser}}
   \vspace{5mm} 
}]

\begin{abstract}
    We present a new approach to the problem of estimating the 3D room layout from a single panoramic image. We represent room layout as three 1D vectors that encode, at each image column, the boundary positions of floor-wall and ceiling-wall, and the existence of wall-wall boundary. The proposed network, \modelname, trained for predicting 1D layout, outperforms previous state-of-the-art approaches. The designed post-processing procedure for recovering 3D room layouts from 1D predictions can automatically infer the room shape with low computation cost---it takes less than 20ms for a panorama image while prior works might need dozens of seconds. We also propose Pano Stretch Data Augmentation, which can diversify panorama data and be applied to other panorama-related learning tasks. Due to the limited data available for non-cuboid layout, we re-label 65 general layout from the current dataset for fine-tuning. Our approach shows good performance on general layouts by qualitative results and cross-validation.
\end{abstract}

\section{Introduction}

The goal of this work is to predict the room layout from a panoramic image. Most of the state-of-the-art methods solve this problem by adopting more effective deep network architectures for their models to learn from different cues in the image. Assumptions about the room structures are often made to constrain the solution space so that the predictions of the deep model would not deviate from the common cases too much. Post-processing steps can further be performed to refine the predictions. Given a number of images with annotated layouts for training, state-of-the-art methods are able to achieve good results on the test data. However, acquiring high-quality room-layout annotations for panoramic images is labor-demanding. The annotations done by different people might be inconsistent due to ambiguities about the locations of wall boundaries, especially for well-decorated rooms. 
Moreover, currently available datasets do not include more images of complex room layouts. The annotation for a complex layout would just be approximated as a cuboid-shaped or L-shaped layout, introducing even more ambiguities for training and testing. 

Two important and correlated issues may be further addressed for improving state-of-the-art methods. The first issue is the lack of more training and validation data with precise annotations. The second issue is that, without more annotated data for training, the deep networks cannot be too large, otherwise the test accuracy might be low due to overfitting.
Collecting more data to train a more sophisticated model is indeed beneficial and doable, but a more efficient way to improve the performance should also be welcome. We argue that, if we have some better understanding of the problem and make good use of domain knowledge, we may improve the performance without acquiring a lot more annotated data or using a larger deep network. Data augmentation is a common procedure in deep learning to generate more data for training. Standard data augmentation heuristics such as random cropping or luminance change for image classification or object detection might not be effective for layout prediction. Our idea is to take account of the underlying geometric constraints and design a better data augmentation mechanism specifically for training layout-predicting deep networks. On the other hand, instead of increasing the model complexity, we aim to enhance the model by devising a compact representation with respect to the geometric constraints. We can, therefore, remove redundant degrees of freedom and force the model to focus more on learning critical properties for layout prediction. 

We characterize our contributions as follows:
\begin{itemize}
    \item We introduce a 1D $\mathcal{O}(W)$ representation that encodes the whole-room layout for a panoramic scene. Training with such a representation allows our method to outperform previous state-of-the-art results, yet requires fewer parameters and less computation time.
    \item We propose a data augmentation mechanism called \textit{Pano Stretch Data Augmentation}, which generates panorama images on the fly during training and improves the accuracy under all settings in our experiments. This data augmentation mechanism also has the potential for boosting other tasks (\eg, semantic segmentation, object detection) that directly work on a panorama.
    \item We show that leveraging RNNs in a layout prediction task is helpful for improving the accuracy. RNNs are able to capture the long-range geometric pattern of room layouts.
    \item Owing to the 1D representation and our efficient post-processing procedure, the computation cost of our model is very low, and the model can be easily extended to handle complex scenes with layouts other than cuboid-shaped or L-shaped.
\end{itemize}

\noindent Code and data are available at: \url{https://sunset1995.github.io/HorizonNet/}.

\section{Related Work}

Room layout estimation from a single-view RGB image is an active research topic over the past decade. Many approaches have been developed in this field. Most of them exploit the Manhattan world assumption that the room layouts, and even the furniture, are aligned with the three principal axes~\cite{coughlan1999manhattan}. The Manhattan world assumption imposes constraints on the layout estimation problem, and, based on the assumption, the Manhattan aligned vanishing points could also be used to rectify the image and extract features for inferring the layout.

Delage \etal~\cite{delage2006dynamic} train a dynamic Bayesian network to recognize the floor-wall boundary in each column of the perspective image. Many approaches search the Manhattan aligned layout based on extracted geometric cues. Lee \etal~\cite{lee2009geometric} test the hypothesis using Orientation Map (OM) while Hedau \etal~\cite{hedau2009recovering} using Geometric Context (GC)~\cite{hoiem2007recovering}. Hedau \etal~\cite{gupta2010estimating} further jointly inference the room layout with 3D objects, \eg beds. Similar strategies have also been used by later methods, such as introducing an improved scoring function~\cite{schwing2012efficient,urtasun2012efficient}, generating layout hypothesis with Manhattan junction~\cite{ramalingam2013manhattan}, and modeling the interaction between objects and layout~\cite{del2013understanding,gupta2010estimating,zhao2013scene}.

The aforementioned methods only deal with perspective images. Zhang \etal~\cite{zhang2014panocontext} propose to estimate the layout from a \threesixty H-FOV panoramic image. They extend the previous methods of vanishing point detection, hypothesis generation, and scoring hypotheses based on OM, GC and object interaction, and apply all of them to panoramas. Xu \etal~\cite{xu2017pano2cad} also use the OM, GC, object detection, and object orientation to reconstruct 3D layout. Yang \etal~\cite{yang2016efficient} use superpixels and Manhattan aligned line segments as features, and formulate the problem by constraint graphs. The method of \cite{yang2018automatic} follows a similar approach using more geometric and semantic features. Other approaches attempt to recover the floor plan from a panorama using image gradient cues~\cite{pintore2016omnidirectional} or from multiple panorama images~\cite{cabral2014piecewise}.

Recent methods rely more on deep networks to improve layout estimation.
Most of them leverage dense prediction models to classify geometric or semantic label for each pixel.
For perspective images, common ways are to predict the boundary probability map~\cite{mallya2015learning,ren2016coarse}, classes of boundaries~\cite{zhao2017physics,ren2016coarse}, classes of layout surface~\cite{dasgupta2016delay,izadinia2017im2cad}, and corner keypoints heatmaps~\cite{lee2017roomnet}.
The predicted dense maps can be post-processed to generate layouts.
A few deep learning methods have been developed for panorama-based layout estimation.
Zou \etal~\cite{zou2018layoutnet} predict the corner probability map and boundary map directly from a panorama.
They also extend Stanford 2D-3D dataset~\cite{2017arXiv170201105A} with annotated layouts for training and evaluation.
Fernandez-Labrador \etal~\cite{fernandez2018layouts} train the deep network on perspective images.
During testing, they stitch the predicted perspective boundary maps into a panorama and combine them with geometric cues to infer the layout.
\revise{
Two concurrent works DuLa-Net~\cite{yang2018dula} and CFL~\cite{fernandez2019CFL} show improved quantitative results with the ability to produce general room shape not limited to cuboid shape.
DuLa-Net~\cite{yang2018dula} combines the surface semantic mask from conventional equirectangular view and the projected floor and ceiling view.
CFL~\cite{fernandez2019CFL} proposes convolution kernel specialized for equirectangular image.
}

Unlike all the \revise{existing} methods that use neural networks to perform dense prediction for layout estimation, we leverage the property of aligned panorama image to predict the positions of floor-wall and ceiling-wall boundaries, as well as the existence of wall-wall boundary for each column of an equirectangular image.
Our model only produces three values for each column of an image, and thus the output size of the model is reduced from $\mathcal{O}(HW)$ to $\mathcal{O}(W)$.
The proposed output representation is similar to~\cite{delage2006dynamic} but they only predict floor-wall boundary for each column of a perspective image using a Dynamic Bayesian Network.
In contrast, our work can handle panoramas and recognize floor-wall, ceiling-wall and wall-wall boundaries using a deep neural network.  
\revise{Existing works~\cite{zou2018layoutnet,fernandez2018layouts,yang2018dula,fernandez2019CFL} on the same task learn to make dense $\mathcal{O}(HW)$ predictions over the entire image while our model predicts only three values for each image column.}
RoomNet~\cite{lee2017roomnet} imitates RNN's recurrent structure with ``time steps" equal to refinement steps. We use RNN where each ``time step" is responsible for estimating the result across a few image columns.

\section{Approach}

\begin{figure*}[h]
   \centering
   \setlength\tabcolsep{1pt}
\includegraphics[width=0.96 \linewidth]{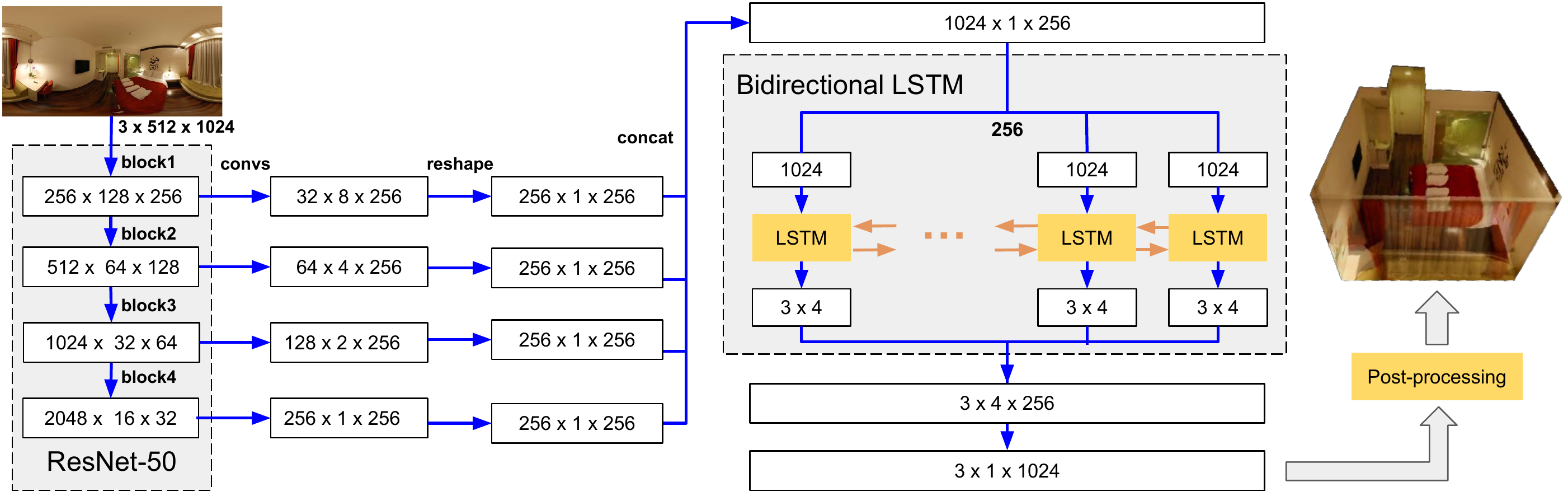}
    \caption{An illustration of the \modelname architecture. }
    \label{fig:model_arch}
\end{figure*}

The goal of our approach is to estimate Manhattan room layout from a panoramic image that covers \threesixty H-FOV. Unlike conventional dense prediction (target output size $=\mathcal{O}(HW)$) for layout estimation using deep learning~\cite{dasgupta2016delay,fernandez2018layouts,fernandez2018panoroom,izadinia2017im2cad,mallya2015learning,ren2016coarse,zhao2017physics}, we formulate the problem as regressing the boundaries and classifying the corner for each column of image (target output size $=\mathcal{O}(W)$). The proposed \modelname trained for predicting the $\mathcal{O}(W)$ target is presented in Sec.~\ref{sssec:approach_net}. In Sec.~\ref{sssec:approach_post}, we introduce a simple yet fast and effective post-processing procedure to derive the layout from output of \modelname. Finally in Sec.~\ref{sssec:approach_aug}, we introduce \textit{Pano Stretch Data Augmentation} which effectively augments the training data on-the-fly by stretching the image and ground-truth layout along \textit{x} or \textit{z} axis (Fig.~\ref{fig:stretch_aug_demo}).

All training and test images are pre-processed by the panoramic image alignment algorithm mentioned in~\cite{zou2018layoutnet}. Our approach exploits the properties of the aligned panoramas that the wall-wall boundaries are vertical lines under equirectangular projection. Therefore, we can use only one value to indicate the column position of wall-wall boundary instead of two (each for a boundary endpoint).

\subsection{\modelname} \label{sssec:approach_net}

Fig.~\ref{fig:model_arch} shows an overview of our network, which comprises a feature extractor and a recurrent neural network. The network takes a single panorama image with the dimension of $3 \times 512 \times 1024$ (channel, height, width) as input.

\begin{figure}[h]
   \centering
   \setlength\tabcolsep{1pt}
\includegraphics[width=0.98 \linewidth]{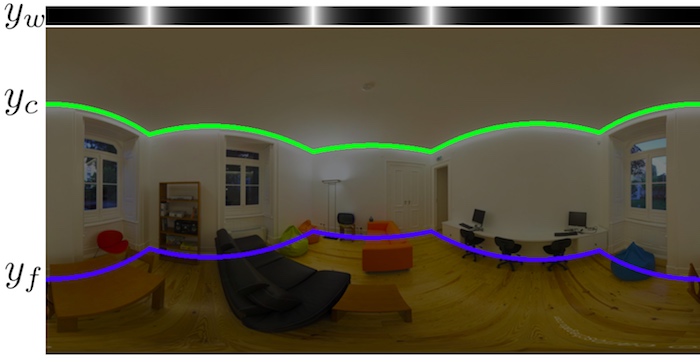}
    \caption{Visualization of our 1D ground truth representations. $y_w$ denotes the existence probability of wall-wall boundary. $y_c, y_f$ (plotted in green and blue) denote the positions of the ceiling-wall boundary and floor-wall boundary respectively. For better visualization, we plot $y_w, y_c, y_f$ with line width greater than one pixel.}
    \label{fig:gt}
\end{figure}

\paragraph{1D Layout Representation:} ~ The size of network output is $3 \times 1 \times 1024$.
As illustrated in Fig.~\ref{fig:gt}, two of the three output channels represent the ceiling-wall ($y_c$) and the floor-wall ($y_f$) boundary position of each image column, and the other one ($y_w$) represents the existence of wall-wall boundary (\ie corner). The values of $y_c$ and $y_f$ are normalized to $[-\pi/2, \pi/2]$.
Since defining $y_w$ as a binary-valued vector with 0/1 labels would make it too sparse to detect (only 4 out of 1024 non-zero values for simple cuboid layout), we set $y_w(i) = c^{dx}$ where $i$ indicates the $i$th column, $dx$ is the distance from the $i$th column to the nearest column where wall-wall boundary exists, and $c$ is a constant.
\revise{To check the robustness of our method against the choice of $c$, we have tried $0.6, 0.8, 0.9, 0.96, 0.99$ and get similar results. Therefore, we stick to $c=0.96$ for all the experiments.}
One benefit of using 1D representation is that it is less affected by zero dominant backgrounds.
2D whole-image representations of boundaries and corners would result in 95\% zero values even after smoothing~\cite{zou2018layoutnet}.
Our 1D boundaries representation introduces no zero backgrounds because the prediction for each component of $y_c$ or $y_f$ is simply a real-valued regression to the ground truth.
The 1D wall-wall (corners) representation also changes the peak-background ratio of ground truth from $\frac{2N}{512 \cdot 1024}$ to $\frac{N}{1024}$ where $N$ is the number of wall-wall corners.
Therefore, the 1D wall-wall representation is also less affected by zero-dominated background.
In addition, computation of 1D compact output is more efficient compared to 2D whole-image output.
As depicted in Sec.~\ref{sssec:approach_post}, recovering the layout from our three 1D representations is simple, fast, and effective.


\paragraph{Feature Extractor:} ~ We adopt ResNet-50 \cite{HeZRS16} as our feature extractor. The output of each block of ResNet-50 has half spatial resolution compared to that of the previous block. To capture both low-level and high-level features, each block of the ResNet-50 contains a sequence of convolution layers in which the number of channels and the height is reduced by a factor of $8$ ($=2\times 2\times 2$) and $16$ ($=4\times2\times2$), respectively. More specifically, each block contains three convolution layers with $4 \times 1, 2 \times 1, 2 \times 1$ kernel size and stride, and the number of channels after each Conv is reduced by a factor of $2$. All the extracted features from each layer are upsampled to the same width $256$ (a quarter of input image width) and reshaped to the same height. The final concatenated feature map is of size  $1024 \times 1 \times  256$. The activation function after each Conv is ReLU except the final layer in which we use Sigmoid for $y_w$ and an identity function for $y_c, y_f$. We have tried various settings for the feature extractor, including deeper ResNet-101, different designs of the convolution layers after each ResNet block, and upsampling to the image width $1024$, and find that the results are similar. Therefore, we stick to the simpler and computationally efficient setting.

\paragraph{Recurrent Neural Network for Capturing Global Information:} ~ Recurrent neural networks (RNNs) are capable of learning patterns and long-term dependencies from sequential data. Geometrically speaking, any corner of a room can be roughly inferred from the positions of other corners; therefore, we use the capability of RNN to capture global information and long-term dependencies. Intuitively, because LSTM~\cite{hochreiter1997long}, a type of RNN architecture, stores information about its prediction for other regions in the cell state, it has the ability to predict for occluded area accurately based on the geometric patterns of the entire room. In our model, RNN is used to predict $y_c', y_f', y_w'$ column by column. That is, the sequence length of RNN is proportional to the image width. In our experiment, RNN predicts for four columns instead of one column per time step, which requires less computational time without loss of accuracy. As the $y_c, y_f, y_w$ of a column is related to both its left and right neighbors, we adopt the bidirectional RNN~\cite{schuster1997bidirectional} to capture the information from both sides. Fig.~\ref{fig:visualize_rnn} and Table~\ref{table:quan_pano} demonstrate the difference between models with or without RNN.

\subsection{Post-processing} \label{sssec:approach_post}


We recover general room layouts that are not limited to cuboid under following assumptions: \emph{i}) intersecting walls are perpendicular to each other (Manhattan world assumption); \emph{ii}) all rooms have the one-floor-one-ceiling layout where floor and ceiling are parallel to each other; \emph{iii}) camera height is 1.6 meters following~\cite{zhang2014panocontext}; \emph{iv}) the pre-processing step correctly align the floor orthogonal to y-axis.

\revise{
As described in Sec.~\ref{sssec:approach_net}, raw outputs of our deep model $y_f', y_c', y_w' \in \mathcal{R}^{1024}$ contain the layout information for each image column. 
Each value in $y_f'$ and $y_c'$ is the position of floor-wall boundary and ceiling-wall boundary at the corresponding image column.
$y_w'$ represents the probability of wall-wall existence of each image column.
}

\paragraph{Recovering the Floor and Ceiling Planes:} ~
\revise{
For each column of the image, we can use the corresponding values in $y_f', y_c'$ to vote for the ceiling-floor distance.
Based on the assumed camera height, we can project the floor-wall boundary $y_f'$ from image to 3D $XYZ$ position (they all shared the same $Y$).
The ceiling-wall boundary $y_c'$ shares the same 3D $X, Z$ position with the $y_f'$ on the same image column, and therefore the distance between floor and ceiling can be calculated.
We take the average of results calculated from all image columns as the final floor-ceiling distance.
}

\paragraph{Recovering Wall Planes:} ~
\revise{We first find the prominent peaks on the estimated wall-wall probability $y_w'$ with two criteria: \emph{i}) the signal should be larger than any other signal within 5\textdegree H-FOV, and \emph{ii}) the signal should be larger than 0.05.}

\begin{figure}[h]
    \centering
    \begin{subfigure}[b]{0.96\linewidth}
        \centering
        \includegraphics[width=\textwidth]{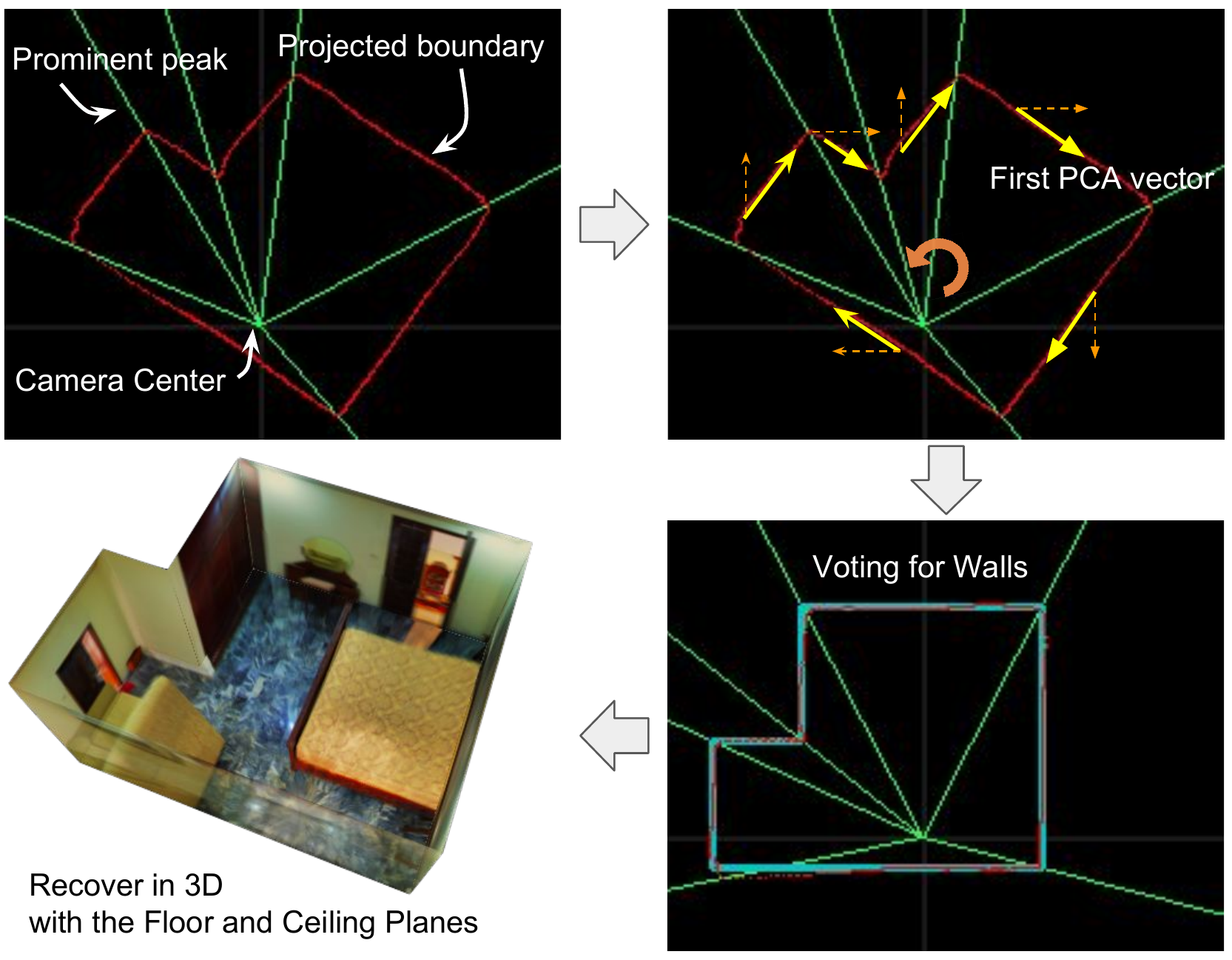}
        \caption{Depicting how we recover the wall planes from our model output.}
        \label{fig:post_walls_a}
    \end{subfigure}
    \begin{subfigure}[b]{0.96\linewidth}
        \centering
        \includegraphics[width=\textwidth]{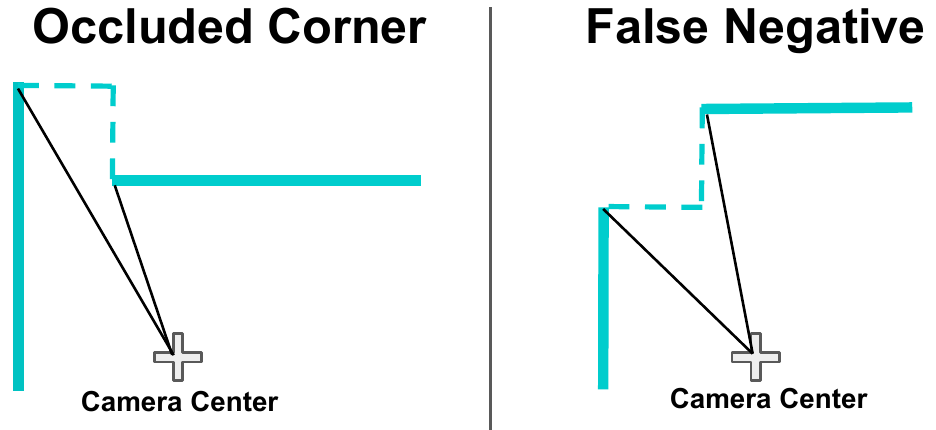}
        \caption{Two special cases: Instead of voting for a wall, we add a corner according to the two prominent peaks and the positions of two walls.}
        \label{fig:post_walls_b}
    \end{subfigure}
    \caption{Visualization of wall planes recovering. Fig.~\ref{fig:post_walls_a} is an example that the pre-processing algorithm fails to correctly align the horizontal rotation of panorama.}
    \label{fig:post_walls}
\end{figure}

Fig.~\ref{fig:post_walls_a} shows the projected $y_c'$ (red points) on ceiling plane.
The green lines are the detected prominent peaks which split the ceiling-wall boundary (red points) into multiple parts.
\revise{
To handle possibly failed horizontal alignment in the pre-processing step, we calculate the first principal component of each part, then rotate the scene by the average angle of all first principal components (top right figure in Fig.~\ref{fig:post_walls_a}).
}
\revise{So now we have two types of walls: \emph{i}) X-axis orthogonal walls and \emph{ii}) Z-axis orthogonal walls.
We construct the walls from low to high variance suggested by the first principal component.
Adjacency walls are forced to be orthogonal to each other, thus only walls whose two adjacent walls are not yet constructed have the freedom to decide the orthogonal type.
}
We use a simple voting strategy: each projected red point votes for all planes within 0.16 meters (bottom right figure in Fig.~\ref{fig:post_walls_a}).
The most voted plane is selected.
Two special cases are depicted in Fig~\ref{fig:post_walls_b} which occur when the two adjacency walls are already constructed and they are orthogonal to each other.
\revise{Finally, the $XYZ$ positions of all corners are decided according to the intersection of three adjacent Manhattan junction planes.}

The time complexity of our post-processing procedure is $\mathcal{O}(W)$, where $W$ is the image width. Thus the post-processing can be efficiently done; in average, it takes less than 20ms to finish.

\subsection{Pano Stretch Data Augmentation} \label{sssec:approach_aug}

\begin{figure}
   \centering
   \setlength\tabcolsep{1pt}
\begin{tabular}{cc}
\makecell{$k_x=1.0, k_z=1.0$  (original)\\\includegraphics[width=0.48\linewidth]{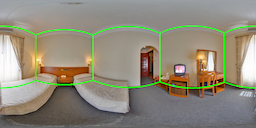}}&
\makecell{$k_x=2.0, k_z=1.0$\\\includegraphics[width=0.48\linewidth]{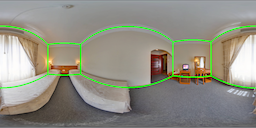}}\\
\makecell{$k_x=1.0, k_z=2.0$\\\includegraphics[width=0.48\linewidth]{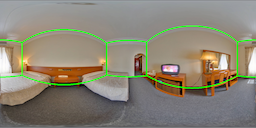}}&
\makecell{$k_x=2.0, k_z=2.0$\\\includegraphics[width=0.48\linewidth]{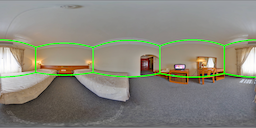}}\\
\end{tabular}

    \caption{Visualization of the proposed \textit{Pano Stretch Data Augmentation}. The image and ground-truth layout (green lines) are stretched along $x$ or $z$ axis (the effect of scaling $y$ can be covered by $x$ and $z$). This can augment the data by changing the room's length and width. This augmentation strategy improves our quantitative results under all experiment settings (Table~\ref{table:ablation}).
    }
    \label{fig:stretch_aug_demo}
\end{figure}

For a \threesixty H-FOV panoramic image, we propose to stretch along axes in 3D space to augment training data. To achieve this goal, we first represent each pixel under UV space as $(u, v)$ where $u \in [-\pi, \pi], v \in [-\pi/2, \pi/2]$. The coordinate $(u, v)$ can be easily computed as the column and row of an equirectangular image, subject to a rotation angle of the camera. Here we introduce an additional variable $d$, which denotes the depth of a pixel. We will show that $d$ can be eliminated later so our final equation does not depend on it.

We project the pixels to 3D space and multiply their $x, y, z$ by $k_x, k_y, k_z$. The equation of stretched $x', y', z'$ are shown in Eq.~\ref{eq:new_xyz}.
\begin{equation} \label{eq:new_xyz}
\left\{\begin{matrix*}[l]
x' = k_x \cdot x = k_x \cdot d \cdot \cos(v) \cdot \cos(u) \,; \\ 
y' = k_y \cdot y = k_y \cdot d \cdot \sin(v) \,; \\ 
z' = k_z \cdot z = k_z \cdot d \cdot \cos(v) \cdot \sin(u) \,.
\end{matrix*}\right.
\end{equation}

We can then project the stretched points back to the sphere by Eq.~\ref{eq:new_uv} for further equirectangular projection. $\atantwo$ in the equation is 2-argument arctangent. The depth $d$ is eliminated since it exists in both terms of $\atantwo$. We fix $k_y = 1$ because setting $k_y$ to a value other than one is equivalent to multiplying $k_x, k_z$ by the same value. 
\begin{equation} \label{eq:new_uv}
\left\{\begin{matrix*}[l]
u' = \atantwo(k_z \cdot \sin(u), ~~ k_x \cdot \cos(u)) \,; \\ 
\begin{aligned}
v' = \atantwo( & k_y \cdot \sin(v), \\
               & \sqrt{k_x^2 \cos^2(u) + k_z^2 \sin^2(u)} \cdot \cos(v) ~~ ) \,.
\end{aligned}
\end{matrix*}\right.
\end{equation}

In our implementation, we do the inverse mapping by Eq.~\ref{eq:ori_uv}. For each pixel in the target image, we compute the corresponding coordinate and sample its value from the source image via bilinear interpolation. Fig.~\ref{fig:stretch_aug_demo} shows a visualization sample.
\begin{equation} \label{eq:ori_uv}
\left\{\begin{matrix*}[l]
u = \atantwo(k_x \cdot \sin(u'), ~~ k_z \cdot \cos(u')) \,; \\ 
\begin{aligned}
v = \arctan(k_z \cdot \tan(v') \cdot \csc(u') \cdot \sin(u)) \,.
\end{aligned}
\end{matrix*}\right.
\end{equation}
Note that our Pano Stretch Data Augmentation procedure could also be used on other tasks (\eg, ground-truth map of semantic segmentation, bounding box for object detection) that directly work on panoramas. The augmentation procedure has the potential to boost the accuracy of those tasks.

\section{Experiments}\label{sssec:exp}

\subsection{Datasets}
We train and evaluate our model using the same dataset as LayoutNet~\cite{zou2018layoutnet}.
The dataset consists of PanoContext dataset~\cite{zhang2014panocontext} and the extended Stanford 2D-3D dataset~\cite{2017arXiv170201105A} annotated by~\cite{zou2018layoutnet}.
To train our model, we generate $3 \times 1 \times 1024$ ground truth from the annotation.
We follow the same training/validation/test split of LayoutNet.

\subsection{Training Details}
The Adam optimizer~\cite{adam} is employed to train the network for 300 epochs with batch size 24 and learning rate 0.0003.
The L1 Loss is used for the ceiling-wall boundary ($y_c$) and floor-wall boundary ($y_f$).
The Binary Cross-Entropy Loss is used for the wall-wall corner ($y_w$).
The network is implemented in PyTorch~\cite{paszke2017automatic}.
It takes four hours to finish the training on three NVIDIA GTX 1080 Ti GPUs.

The data augmentation techniques we adopt include standard left-right flipping, panoramic horizontal rotation, and luminance change.
Moreover, we exploit the proposed Pano Stretch Data Augmentation (Sec.~\ref{sssec:approach_aug}) during training.
The stretching factors $k_x, k_z$ are sampled from uniform distribution $U[1,2]$, and then take the reciprocals of sampled values with probability $0.5$.
The process time of Pano Stretch Data Augmentation is roughly 130ms per $512 \times 1024$ RGB image.
Therefore, it is feasible to be applied on-the-fly during training.

\subsection{Cuboid Room Results}
We generate cuboid room by only selecting the four most prominent peaks in the post-processing step (Sec.~\ref{sssec:approach_post}).

\paragraph{Quantitative Results:} ~ Our approach is evaluated on three standard metrics: \emph{i}) \textbf{3D IoU}: intersection over union between 3D layout constructed from our prediction and the ground truth; \emph{ii}) \textbf{Corner Error}: average Euclidean distance between predicted corners and ground-truth corners (normalized by image diagonal length); \emph{iii}) \textbf{Pixel Error}: pixel-wise error between predicted surface classes and ground-truth surface classes.

\revise{
The quantitative results of different training and testing settings are summarized in Table~\ref{table:quan_pano} and Table~\ref{table:quan_st2d3d}.
To clarify the difference, the input resolution of DuLa-Net~\cite{yang2018dula} and CFL~\cite{fernandez2019CFL} are $256 \times 512$ while LayoutNet~\cite{zou2018layoutnet} and ours are $512 \times 1024$.
Other than conventional augmentation technique, CFL~\cite{fernandez2019CFL} is trained with Random Erasing while ours is trained with the proposed Pano Stretch.
DuLa-Net~\cite{yang2018dula} did not report corner errors and pixel errors.
Our approach achieves state-of-the-art performance and outperforms existing methods under all settings.
}

\begin{figure*}[h]
   \centering
\setlength\tabcolsep{1.5pt}
\begin{tabular}{cccc}
\makecell{\includegraphics[width=0.24\linewidth]{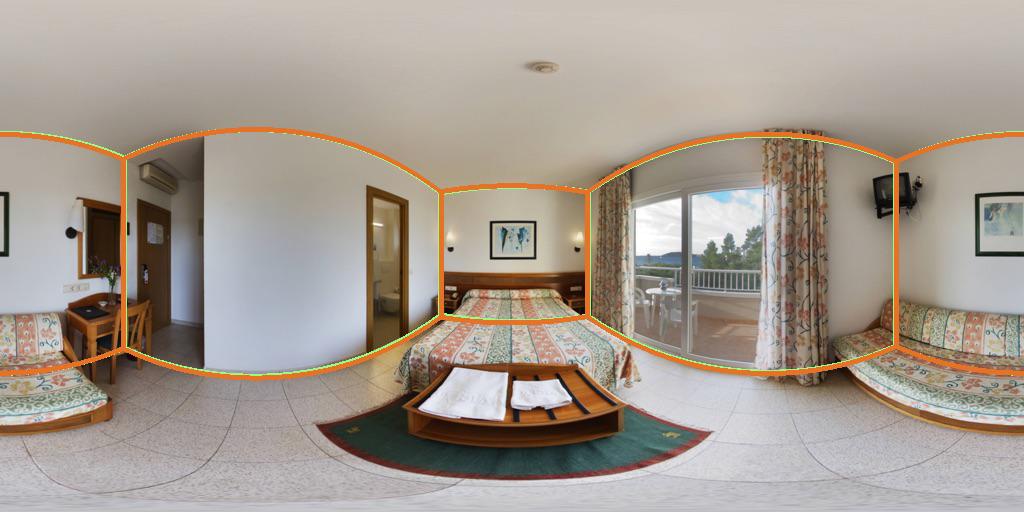}}&
\makecell{\includegraphics[width=0.24\linewidth]{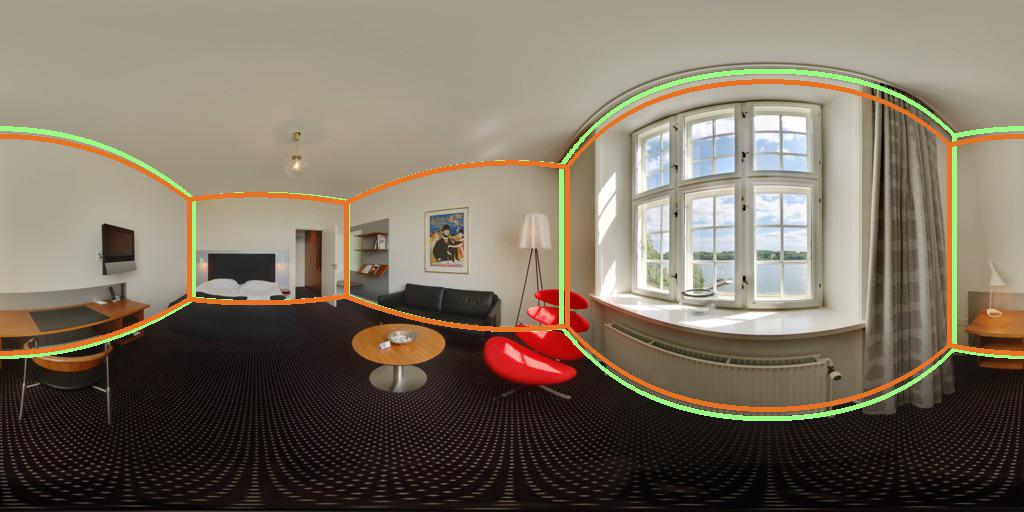}}&
\makecell{\includegraphics[width=0.24\linewidth]{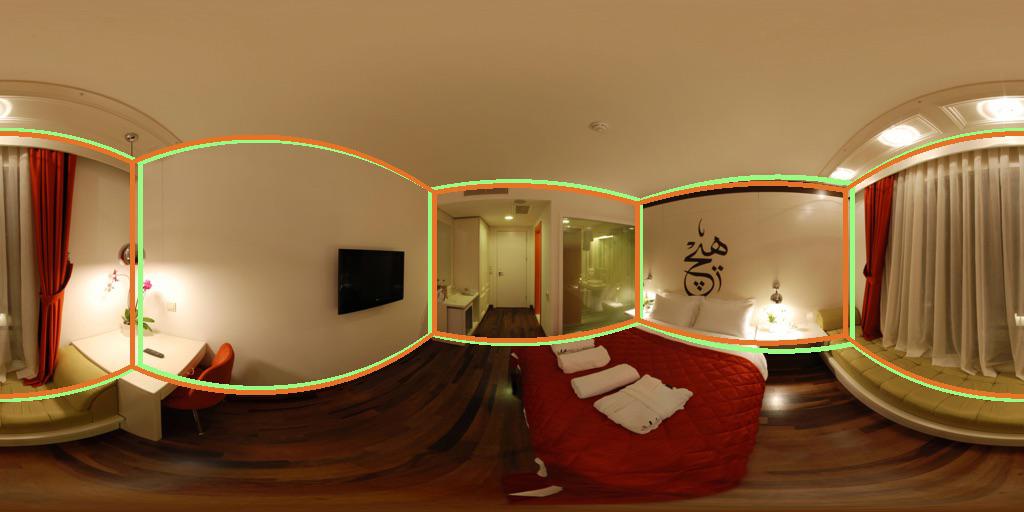}}&
\makecell{\includegraphics[width=0.24\linewidth]{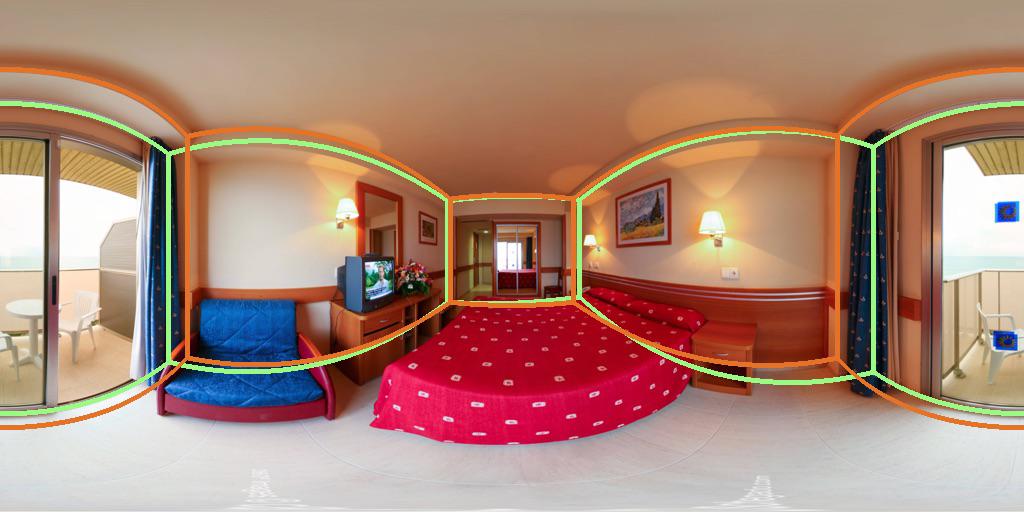}}\\
\makecell{\includegraphics[width=0.24\linewidth]{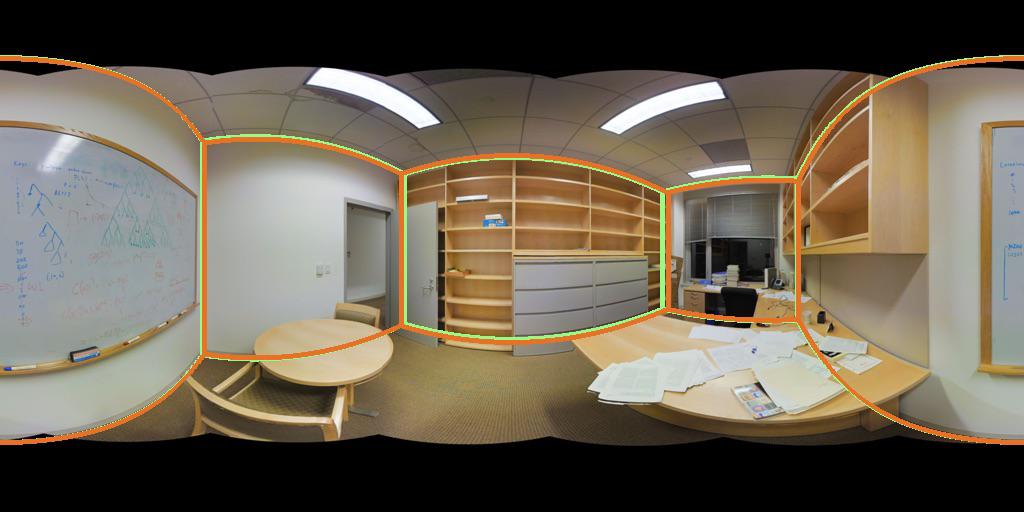}}&
\makecell{\includegraphics[width=0.24\linewidth]{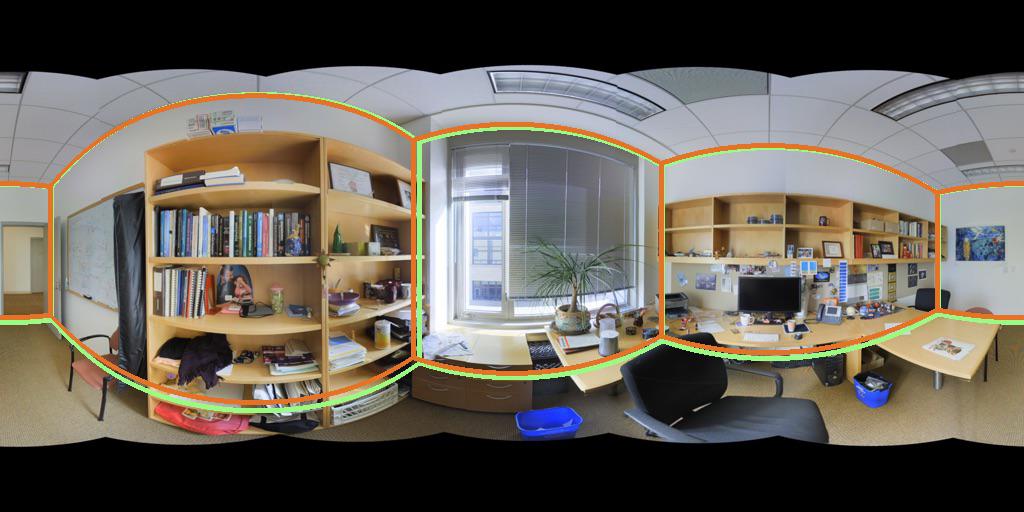}}&
\makecell{\includegraphics[width=0.24\linewidth]{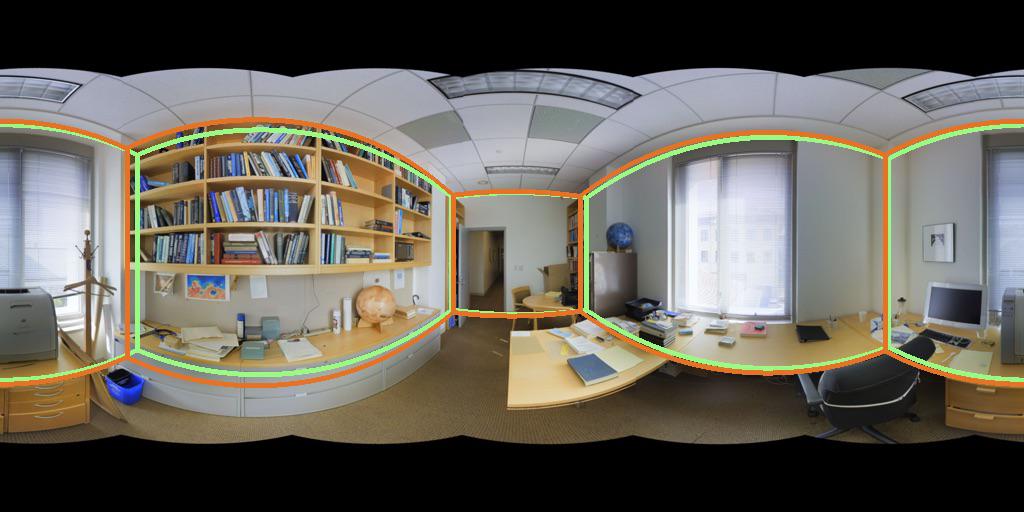}}&
\makecell{\includegraphics[width=0.24\linewidth]{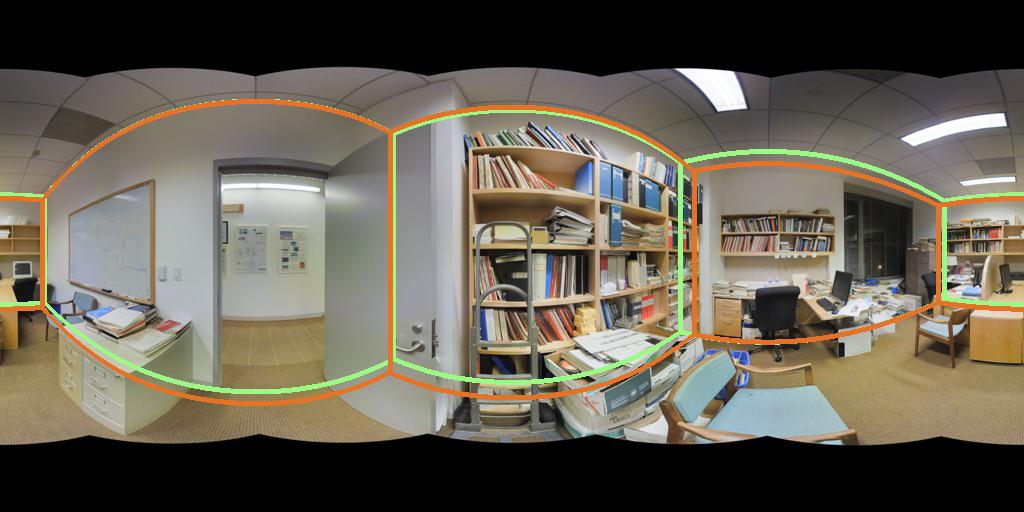}}\\
\end{tabular}
    \caption{
    \revise{
    Qualitative results of cuboid layout estimation.
    The results are separately sampled from four groups that comprise results with the best 0--25\%, 25--50\%, 50--75\% and 75--100\% corner errors (displayed from the first to the fourth columns).
    The green lines are ground truth layout while the orange lines are estimated.
    The images in the first row are from PanoContext dataset~\cite{zhang2014panocontext} while second row are from Stanford 2D-3D dataset~\cite{2017arXiv170201105A}.
    }
    }
    \label{fig:qual_pano}
\end{figure*}

\paragraph{Qualitative Results:} ~
The qualitative results are shown in Fig.~\ref{fig:qual_pano}. We present the results from the best to the worst based on their corner errors. \revise{Please see more results in the supplemental materials.}

\paragraph{Computation time:} ~ The 1D layout representation is easy to compute.
Forward passing a single \textit{512 x 1024} RGB image takes 8ms and 50ms for our \modelname with and without RNN respectively.
The post-processing step for extracting layout from our 1D representation takes only 12ms.
We evaluate the result on a single NVIDIA Titan X GPU and an Intel i7-5820K 3.30GHz CPU.
The reported execution time is averaged across all the testing data.

\begin{table}[t]
    \centering
    \begin{tabular}{|c|c|c|c|} 
        \hline
        Method & 3D IoU(\%) & \makecell{Corner\\error(\%)} & \makecell{Pixel\\error(\%)} \\
        
        \hline\hline
        \multicolumn{4}{|c|}{Train on PanoContext dataset} \\ [0.5ex]
        \hline
        PanoContext~\cite{zhang2014panocontext} & 67.23 & 1.60 & 4.55 \\ 
        \hline
        LayoutNet~\cite{zou2018layoutnet} & 74.48 & 1.06 & 3.34 \\
        \hline
        DuLa-Net~\cite{yang2018dula} & 77.42 & - & - \\
        \hline
        CFL~\cite{fernandez2019CFL} & 78.79 & 0.79 & 2.49 \\
        \hline
        \textbf{ours} & \textbf{82.17} & \textbf{0.76} & \textbf{2.20} \\
        
        \hline\hline
        \multicolumn{4}{|c|}{Train on PanoContext + Stnfd.2D3D datasets} \\ [0.5ex]
        \hline
        LayoutNet~\cite{zou2018layoutnet} &  75.12 &1.02 & 3.18 \\
        \hline
        \textbf{ours} & \textbf{84.23} & \textbf{0.69} & \textbf{1.90} \\
        \hline
    \end{tabular}

    \caption{
    Quantitative results of cuboid layout estimation evaluated on the PaonContext~\cite{zhang2014panocontext} dataset.
    Our method outperforms all existing methods under all settings.
    }
    \label{table:quan_pano}
\end{table}

\begin{table}[h]
    \centering
    \begin{tabular}{|c|c|c|c|} 
        \hline
        Method & 3D IoU(\%) & \makecell{Corner\\error(\%)} & \makecell{Pixel\\error(\%)} \\
        
        \hline\hline
        \multicolumn{4}{|c|}{Train on PanoContext dataset} \\ [0.5ex]
        \hline
        CFL~\cite{fernandez2019CFL} & 65.13 & 1.44 & 4.75 \\
        \hline
        \textbf{ours} & \textbf{75.57} & \textbf{0.94} & \textbf{3.18} \\
        \hline
        
        \hline\hline
        \multicolumn{4}{|c|}{Train on Stnfd.2D3D dataset} \\ [0.5ex]
        \hline
        LayoutNet~\cite{zou2018layoutnet} & 76.33 & 1.04 & 2.70 \\
        \hline
        DuLa-Net~\cite{yang2018dula} & 79.36 & - & - \\
        \hline
        \textbf{ours} & \textbf{79.79} & \textbf{0.71} & \textbf{2.39} \\
        \hline
        
        \hline\hline
        \multicolumn{4}{|c|}{Train on PanoContext + Stnfd.2D3D datasets} \\ [0.5ex]
        \hline
        LayoutNet~\cite{zou2018layoutnet} & 77.51 & 0.92 & 2.42 \\
        \hline
        \textbf{ours} & \textbf{83.51} & \textbf{0.62} & \textbf{1.97} \\
        \hline
        
    \end{tabular}

    \caption{
    Quantitative results of cuboid layout estimation evaluated on the Stanford-2D3D~\cite{2017arXiv170201105A} dataset.
    Our method outperforms all existing methods under all settings.
    }
    \label{table:quan_st2d3d}
\end{table}

\begin{table*}[h]
    \centering
    \begin{tabular}{| >{\centering\arraybackslash} m{2.4cm}  >{\centering\arraybackslash} m{1.8cm}  >{\centering\arraybackslash} m{0.8cm} ||c|c|c||c|c|} 
        \hline
        {Output Shape} & {Stretch Aug.} & RNN & 3D IoU(\%) & {Corner error(\%)} & {Pixel error(\%)} & {\#params} & {FPS} \\
        \hline\hline
        dense $\mathcal{O}(HW)$ &   &   & 77.87 & 1.02 & 2.73 & 67M & 98 \\
        \hline
        dense $\mathcal{O}(HW)$ & V &   & 79.64 & 0.74 & 2.39 & 67M & 98 \\
        \hline
        our $\mathcal{O}(W)$ &   &   & 80.65 & 0.80 & 2.43 & 25M & 119 \\
        \hline
        our $\mathcal{O}(W)$ & V &   & 81.22 & 0.71 & 2.28 & 25M & 119 \\
        \hline
        our $\mathcal{O}(W)$ &   & V & 81.23 & 0.72 & 2.20 & 57M & 20 \\
        \hline
        our $\mathcal{O}(W)$ & V & V & \textbf{83.74} & \textbf{0.65} & \textbf{1.95} & 57M & 20 \\
        \hline
    \end{tabular}

    \caption{
    Ablation study demonstrates the effectiveness of each component in our approach.
    We show that all of our proposed designs can improve the quantitative result.
    Besides, our proposed 1D layout representation significantly reduces the number of parameters.
    FPS is measured for forward-pass of a $3 \times 512 \times 1024$ image on an NVIDIA TITAN X GPU.
    } 
    \label{table:ablation}
\end{table*}

\subsection{Ablation Study}\label{sssec:exp_abla}
Ablation experiments are presented in Table~\ref{table:ablation}. We report the result averaged across all the testing instances.
For a fair comparison, we also experiment with dense $\mathcal{O}(HW)$ prediction following LayoutNet~\cite{zou2018layoutnet} but replace the U-Net~\cite{ronneberger2015u} with the same backbone as our architecture.
\footnote{To output dense (full-image) probability map, we change the Conv layer after each ResNet block from reducing both height and channels to reducing only channels, and then upsample to the same spatial dimension as the input image. Finally, the processed features of four blocks are concatenated and passed through a Conv layer to generate the final result.}
The results of this setting are presented in the first two rows.
We do not try dense $\mathcal{O}(HW)$ output with RNN since it would consume too many computing resources.
We can see that learning on our 1D $\mathcal{O}(W)$ layout representation is better than conventional dense $\mathcal{O}(HW)$ layout representation.

We observe that training with the proposed Pano Stretch Data Augmentation can always boost the performance.
Note that the proposed data augmentation method can also be adopted in other tasks on panoramas and has the potential to increase their accuracy as well.
\revise{See supplemental material for the experiment using Pano Stretch Data Augmentation on semantic segmentation task.}

\begin{figure}
   \centering
\setlength\tabcolsep{0.5pt}
\begin{tabular}{cc}
\makecell{\includegraphics[width=0.48\linewidth]{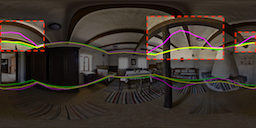}}&
\makecell{\includegraphics[width=0.48\linewidth]{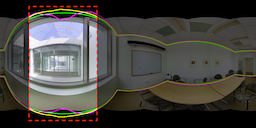}}\\
\makecell{\includegraphics[width=0.48\linewidth]{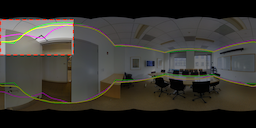}}&
\makecell{\includegraphics[width=0.48\linewidth]{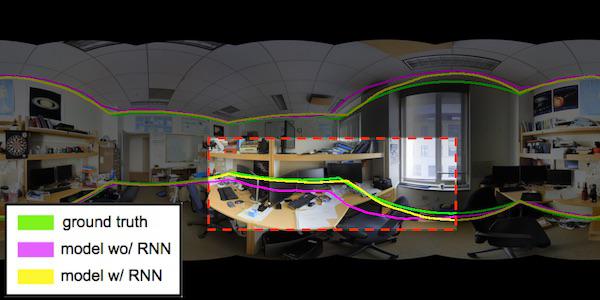}}\\
\end{tabular}
    \caption{
    Visualization of model outputs with and without RNN. We plot the ground truth (green), outputs of the model with RNN (yellow), and outputs of the model without RNN (magenta). Both predictions are raw network outputs without post-processing. The model with RNN performs better than the model without RNN in images contain ceiling beam, black missing polar region caused by smaller camera V-FOV, and occluded area.}
    \label{fig:visualize_rnn}
\end{figure}

For the rows where RNN columns are unchecked, the RNN components shown in Fig~\ref{fig:model_arch} are replaced by fully connected layers. Our experiments show that using RNN in network architecture also improves performance. Fig.~\ref{fig:visualize_rnn} shows some representative results with and without RNN. The raw output of the model with RNN is highly consistent with the Manhattan world even without post-processing, which demonstrates the ability of RNN to capture the geometric pattern of the entire room.

\begin{figure*}
   \centering
\setlength\tabcolsep{2pt}
\begin{tabular}{cccc}
\makecell{\includegraphics[width=0.2\linewidth]{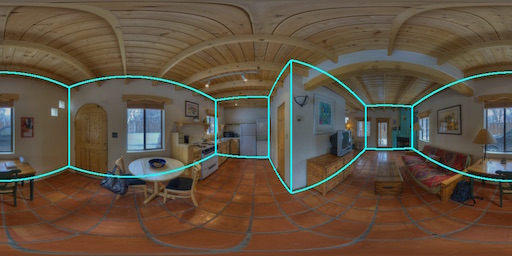}}&
\makecell{\includegraphics[width=0.23\linewidth,height=0.15\linewidth]{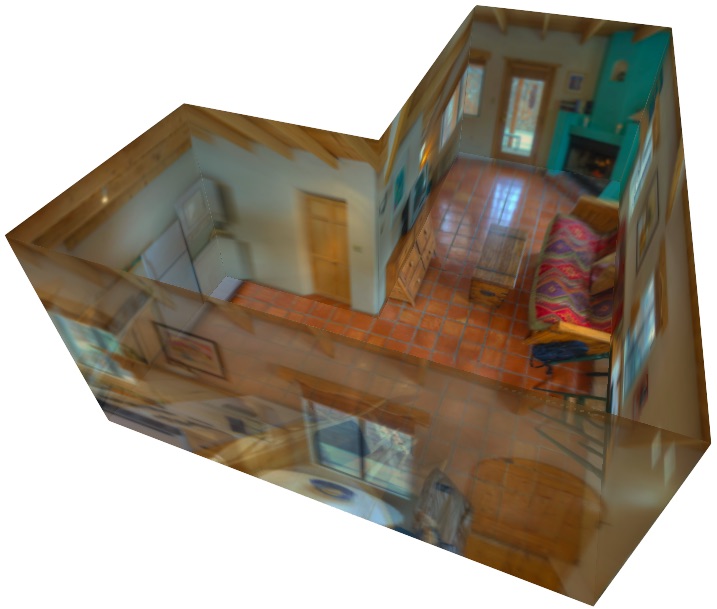}}&
\makecell{\includegraphics[width=0.2\linewidth]{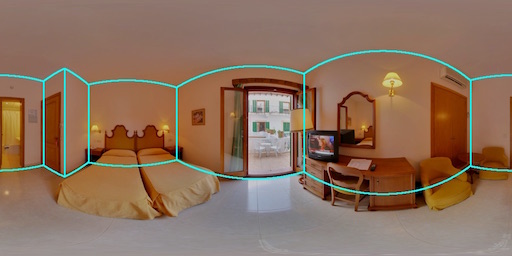}}&
\makecell{\includegraphics[width=0.23\linewidth,height=0.15\linewidth]{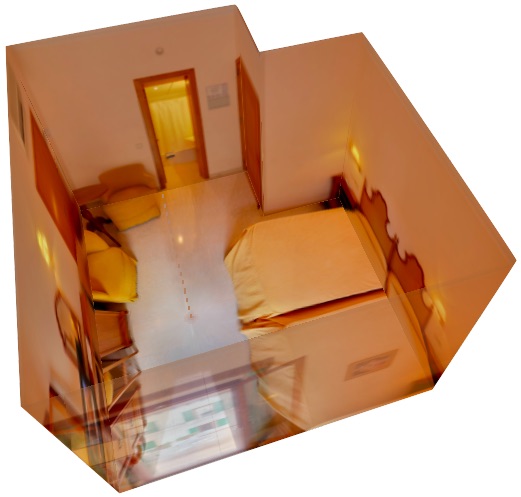}}\\
\makecell{\includegraphics[width=0.2\linewidth]{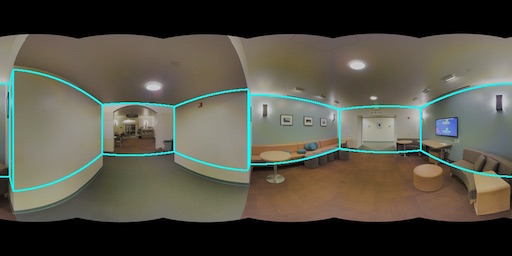}}&
\makecell{\includegraphics[width=0.23\linewidth,height=0.15\linewidth]{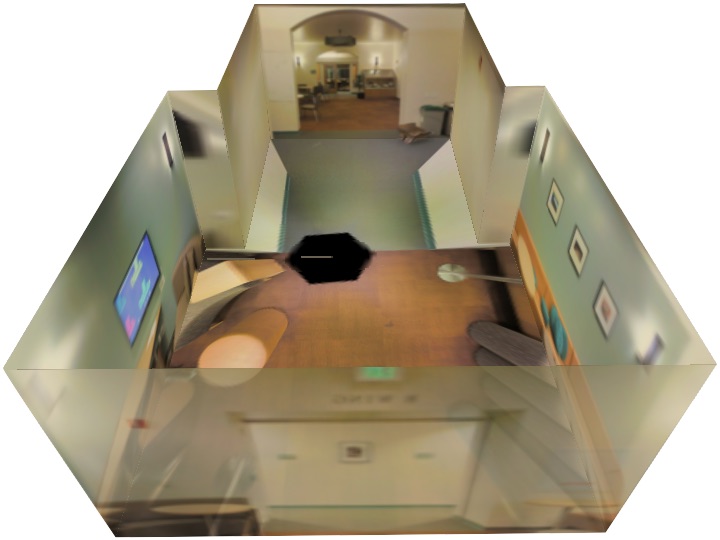}}&
\makecell{\includegraphics[width=0.2\linewidth]{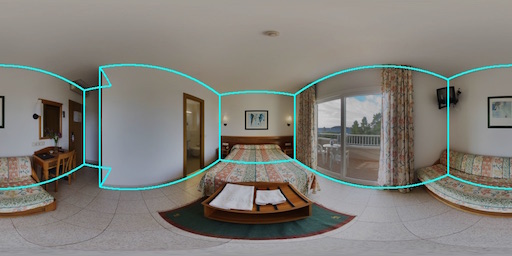}}&
\makecell{\includegraphics[width=0.23\linewidth,height=0.15\linewidth]{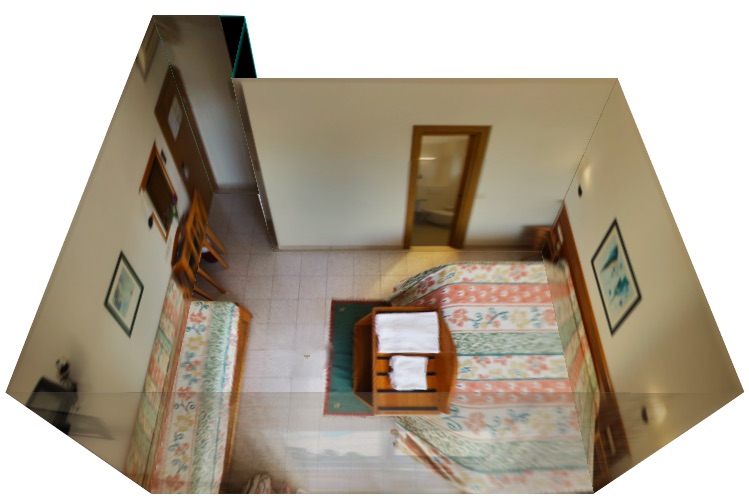}}\\
\makecell{\includegraphics[width=0.2\linewidth]{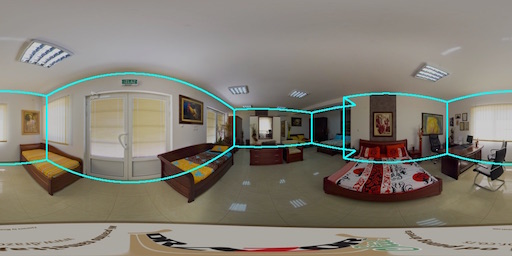}}&
\makecell{\includegraphics[width=0.23\linewidth,height=0.15\linewidth]{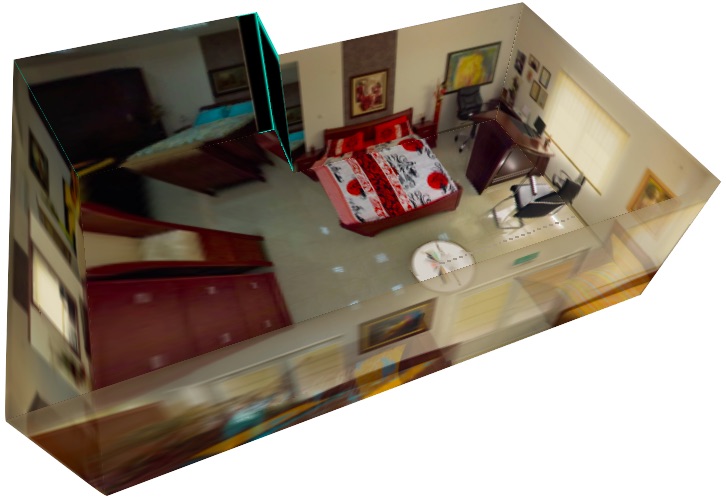}}&
\makecell{\includegraphics[width=0.2\linewidth]{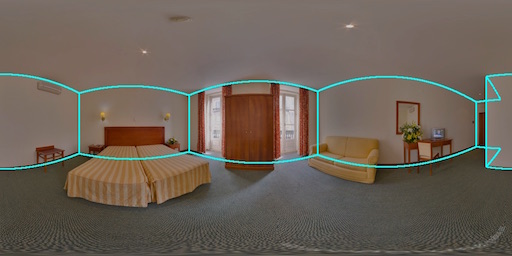}}&
\makecell{\includegraphics[width=0.23\linewidth,height=0.15\linewidth]{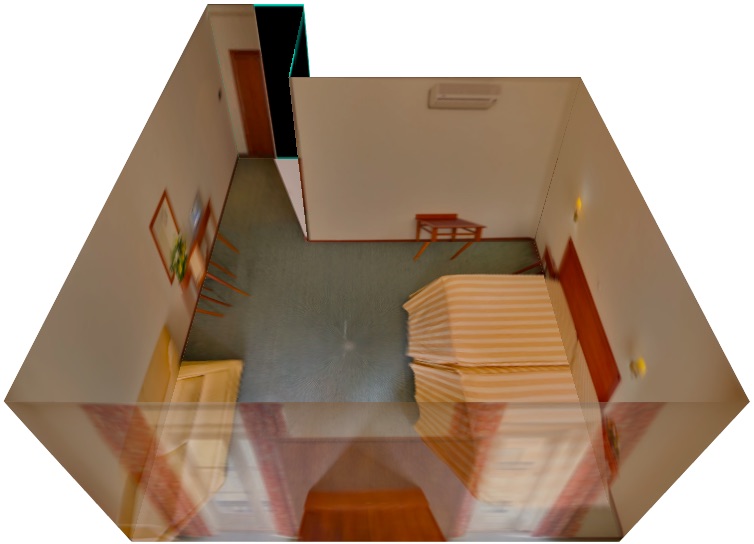}}\\
\end{tabular}
    \caption{
    \revise{
    Qualitative results of non-cuboid layout estimation. The occluded walls are filled with black. The blue lines in the equirectangular images are the estimated room layout boundary.}
    }
    \label{fig:non-cuboid-results}
\end{figure*}

\subsection{Non-cuboid Room Results}\label{sssec:exp_complex}
\revise{
Since the non-cuboid rooms in PanoContext and Stanford 2D-3D dataset are labeled as cuboids, our model is never trained to recognize non-cuboid layouts and concave corners.
This bias makes our model tend to predict complex-shaped rooms as cuboids.
To estimate general room layouts, we re-label 65 rooms from the training split to fine-tune our trained model.
We fine-tune our model for 300 epochs with learning rate $5\mathrm{e}{-5}$ and batch size $2$.
}

\revise{
To quantitatively evaluate the fine-tuning result on general-shaped rooms, we use 13-fold cross validation on the 65 re-annotated non-cuboid data.
The results are summarized in Table~\ref{table:quan_general}.
We depict some examples of reconstructed non-cuboid layouts from the testing and validation splits in Fig.\ref{fig:teaser} and Fig.\ref{fig:non-cuboid-results}.
See supplemental material for more reconstructed layouts.
The results show that our approach can work well on general room layout even with corners occluded by other walls.
}

\begin{table}[h]
    \centering
    \begin{tabular}{|c|c||c|} 
        \hline
        Method & Finetuning & 3D IoU(\%) \\
        \hline\hline
        
        LayoutNet &   & 74.1 \\
        \hline
        LayoutNet & V & 75.1 \\
        \hline
        ours      &   & 77.4 \\
        \hline
        \textbf{ours} & \textbf{V} & \textbf{82.5} \\
        \hline
        
    \end{tabular}

    \caption{
    Quantitative results on the 65 re-annotated non-cuboid datas. The result of fine-tuning is evaluated by 13-fold validation.
    }
    \label{table:quan_general}
\end{table}

\section{Conclusion}
We have presented a new 1D representation for the task of estimating room layout from a panorama. The proposed \modelname trained with such 1D representation outperforms previous state-of-the-art methods and requires fewer computation resources. Our post-processing method which recovers 3D layout from the model output is fast and effective, and it also works for complex room layouts even with occluded corners. The proposed Pano Stretch Data Augmentation further improves our results, and can also be applied to the training procedure of other panorama tasks for potential improvement. 

\vspace{2mm}
\noindent \textbf{Acknowledgement:} This research was partially supported by iStaging and by MOST grants 106-2221-E-007-080-MY3, 107-2218-E-007-047, and 108-2634-F-001-007. 
{\small
\bibliographystyle{ieee}
\bibliography{main}
}

\newpage
\onecolumn
\setcounter{section}{0}
\renewcommand\thesection{\Alph{section}}
\section{Pano Stretch Augmentation for Semantic Segmentation}

We evaluate the potential advantage of the proposed Pano Stretch Data Augmentation on semantic segmentation task. We train and test on Stanford 2D3D~\cite{2017arXiv170201105A} semantic segmentation benchmark. We train PSPNet~\cite{zhao2017pyramid} on subsampled training set and test on the whole testing set. The results are summarized in Table~\ref{table:sem}.

\begin{table}[h]
    \centering
    \begin{tabular}{c|cccccc} 
        \hline
        \# of training images & 20 & 50 & 100 & 200 & 500 & 1040 \\
        \hline
        wo/ pano stretch aug. & 31.5 & 34.9 & 37.1 & 40.7 & 44.2 & 44.8 \\
        w/ pano stretch aug. & \textbf{33.2} & \textbf{36.1} & \textbf{38.4} & \textbf{41.9} & \textbf{44.3} & \textbf{44.9} \\
        \hline
    \end{tabular}

    \caption{
    We evaluate the effect of Pano Stretch Augmentation on semantic segmentation task using the standard metric - mIoU (\%).
    The result implies that the new augmentation technique has the potential to mitigate the "lack of training data" problem for other tasks like semantic segmentation.
    }
    \label{table:sem}
\end{table}

\section{More Qualitative Results of Cuboid Room Layout Reconstruction}

\begin{figure*}[h!]
   \centering
\setlength\tabcolsep{1.5pt}
\begin{tabular}{cccc}
\makecell{\includegraphics[width=0.23\linewidth]{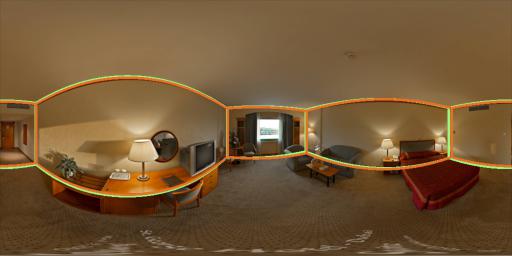}}&
\makecell{\includegraphics[width=0.23\linewidth]{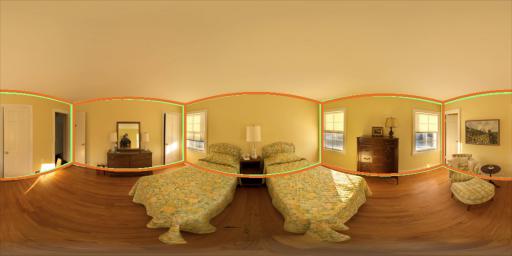}}&
\makecell{\includegraphics[width=0.23\linewidth]{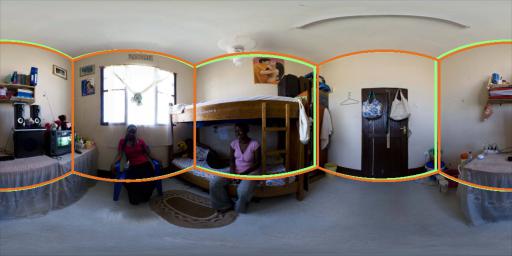}}&
\makecell{\includegraphics[width=0.23\linewidth]{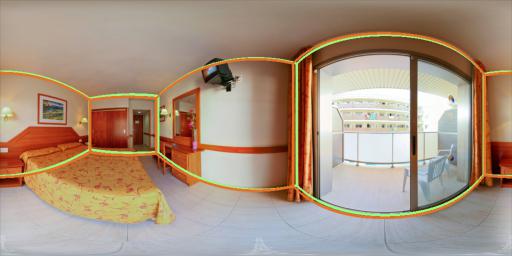}}\\
\makecell{\includegraphics[width=0.23\linewidth]{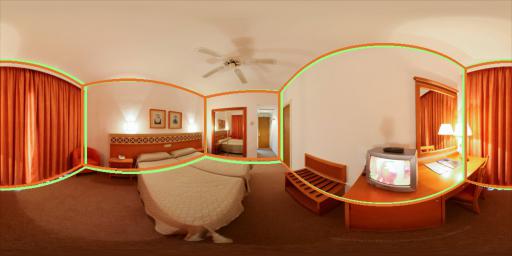}}&
\makecell{\includegraphics[width=0.23\linewidth]{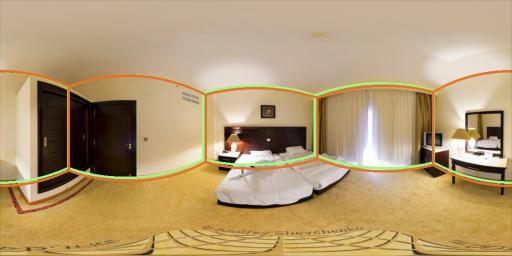}}&
\makecell{\includegraphics[width=0.23\linewidth]{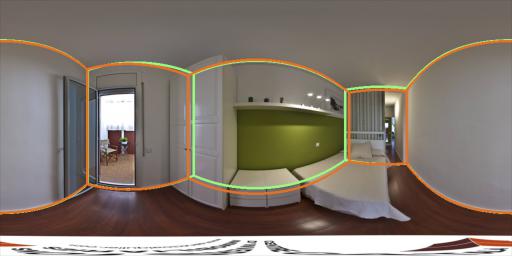}}&
\makecell{\includegraphics[width=0.23\linewidth]{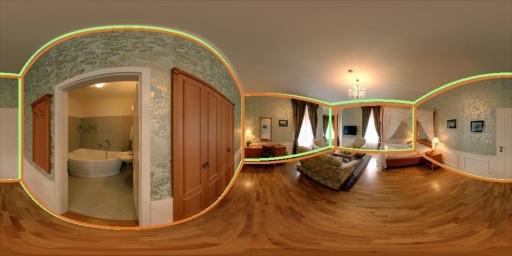}}\\
\makecell{\includegraphics[width=0.23\linewidth]{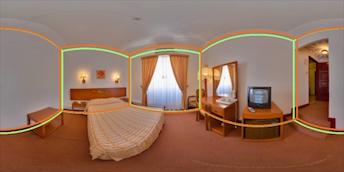}}&
\makecell{\includegraphics[width=0.23\linewidth]{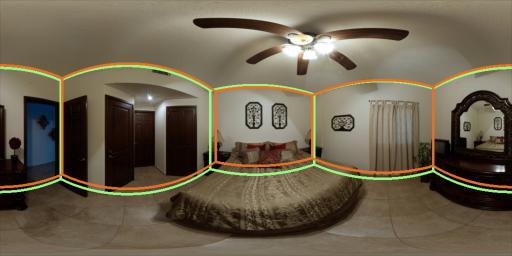}}&
\makecell{\includegraphics[width=0.23\linewidth]{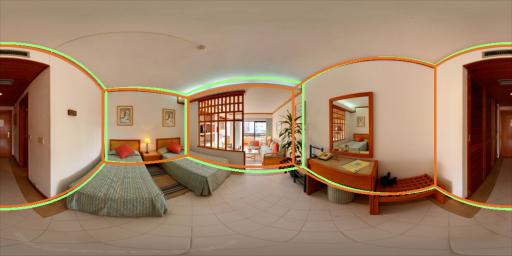}}&
\makecell{\includegraphics[width=0.23\linewidth]{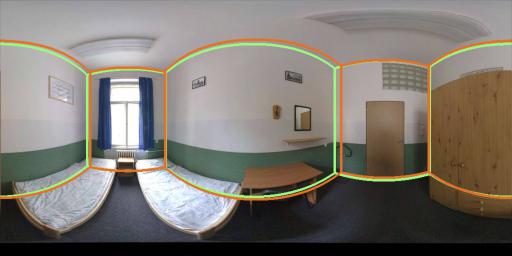}}\\
\makecell{\includegraphics[width=0.23\linewidth]{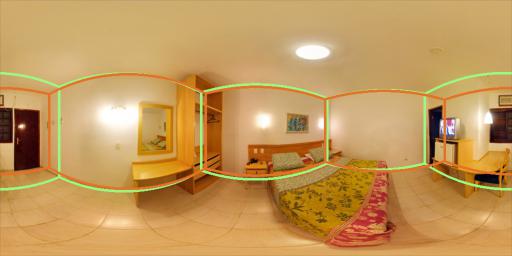}}&
\makecell{\includegraphics[width=0.23\linewidth]{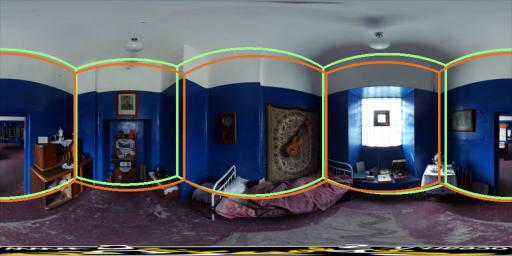}}&
\makecell{\includegraphics[width=0.23\linewidth]{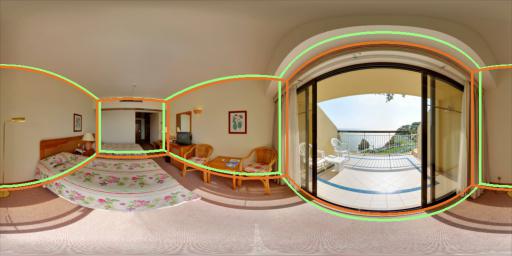}}&
\makecell{\includegraphics[width=0.23\linewidth]{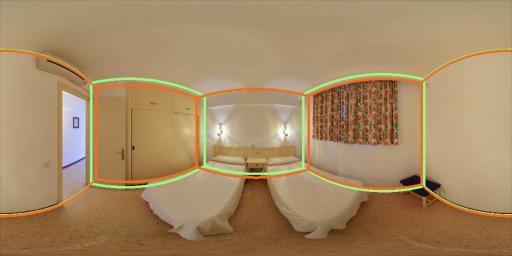}}\\
\makecell{\includegraphics[width=0.23\linewidth]{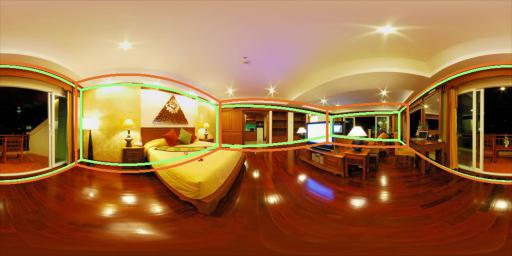}}&
\makecell{\includegraphics[width=0.23\linewidth]{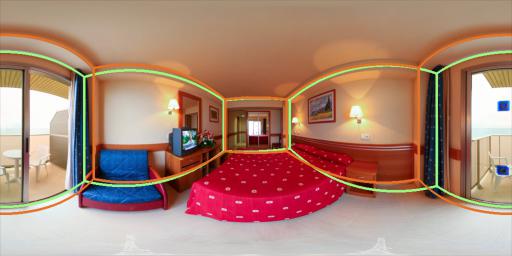}}&
\makecell{\includegraphics[width=0.23\linewidth]{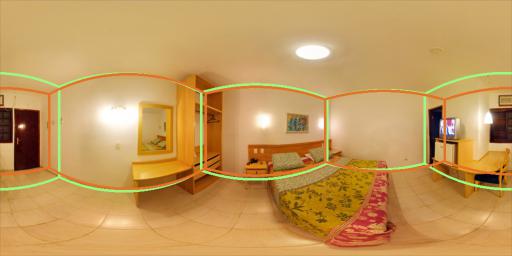}}&
\makecell{\includegraphics[width=0.23\linewidth]{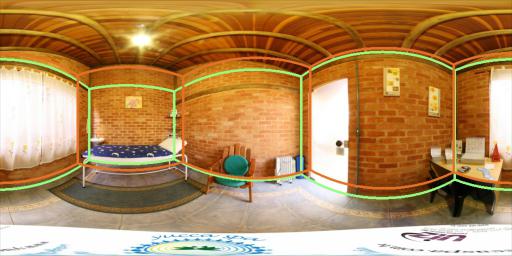}}\\
\end{tabular}
    \caption{
    Qualitative results of cuboid layout estimation on PanoContext~\cite{zhang2014panocontext} dataset.
    The results in the first to the fourth rows are separately sampled from four groups that comprise results with the best 0--25\%, 25--50\%, 50--75\% and 75-100\% corner errors, and the four results with the worst corner errors are displayed in the last row.
    The green lines are ground truth layout while the orange lines are estimated layout.
    }
\end{figure*}

\begin{figure*}[h!]
   \centering
\setlength\tabcolsep{1.5pt}
\begin{tabular}{cccc}
\makecell{\includegraphics[width=0.23\linewidth]{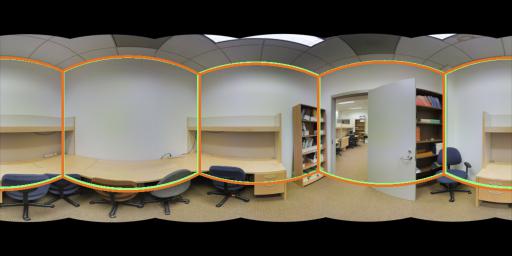}}&
\makecell{\includegraphics[width=0.23\linewidth]{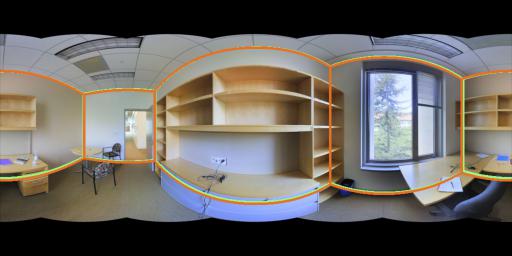}}&
\makecell{\includegraphics[width=0.23\linewidth]{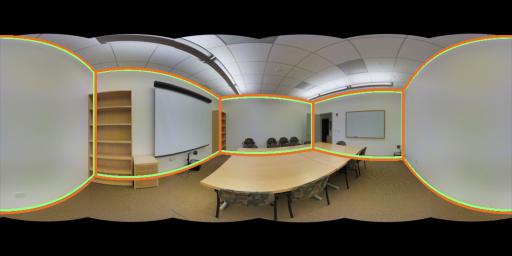}}&
\makecell{\includegraphics[width=0.23\linewidth]{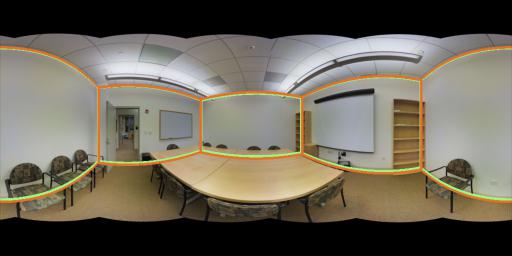}}\\
\makecell{\includegraphics[width=0.23\linewidth]{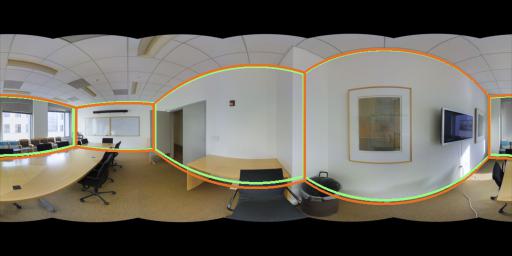}}&
\makecell{\includegraphics[width=0.23\linewidth]{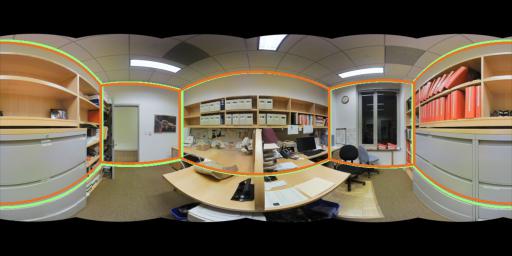}}&
\makecell{\includegraphics[width=0.23\linewidth]{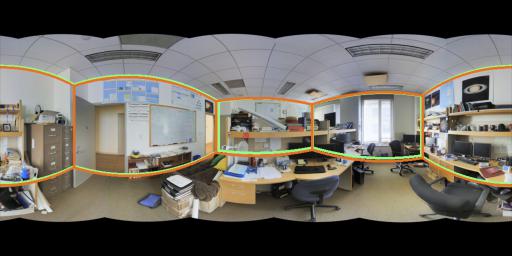}}&
\makecell{\includegraphics[width=0.23\linewidth]{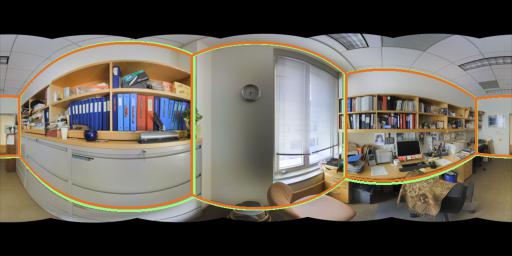}}\\
\makecell{\includegraphics[width=0.23\linewidth]{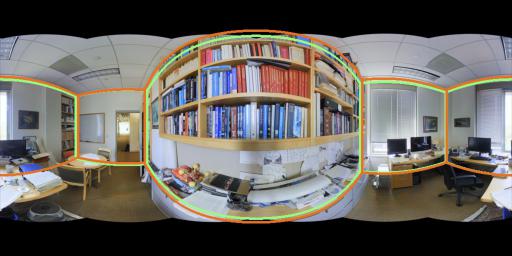}}&
\makecell{\includegraphics[width=0.23\linewidth]{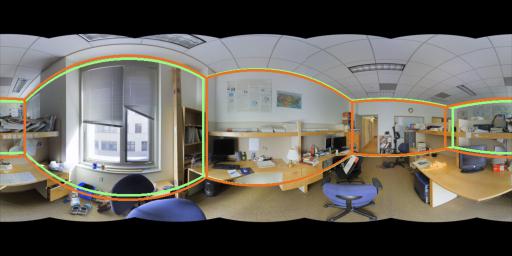}}&
\makecell{\includegraphics[width=0.23\linewidth]{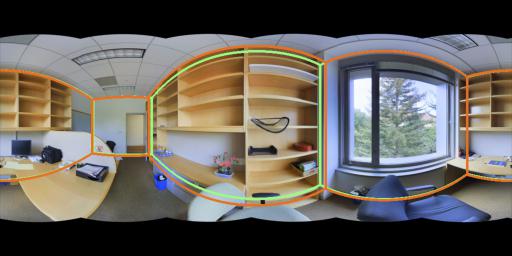}}&
\makecell{\includegraphics[width=0.23\linewidth]{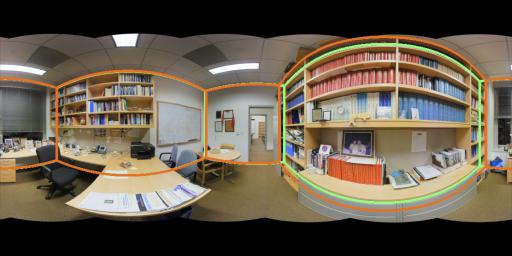}}\\
\makecell{\includegraphics[width=0.23\linewidth]{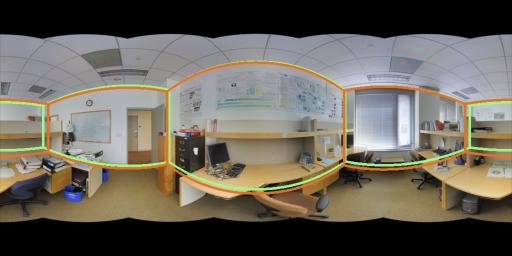}}&
\makecell{\includegraphics[width=0.23\linewidth]{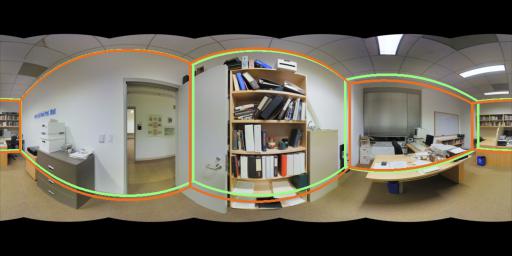}}&
\makecell{\includegraphics[width=0.23\linewidth]{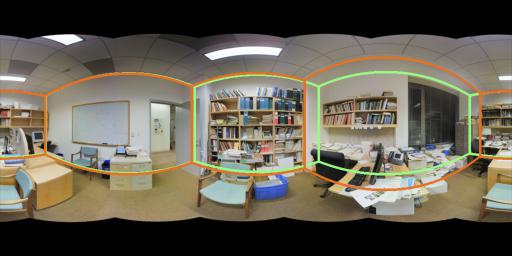}}&
\makecell{\includegraphics[width=0.23\linewidth]{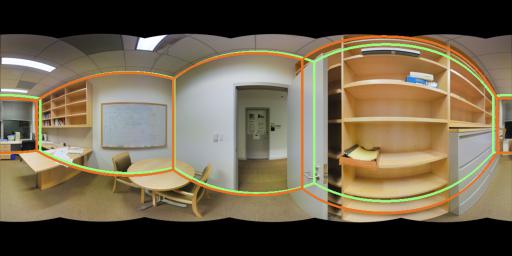}}\\
\makecell{\includegraphics[width=0.23\linewidth]{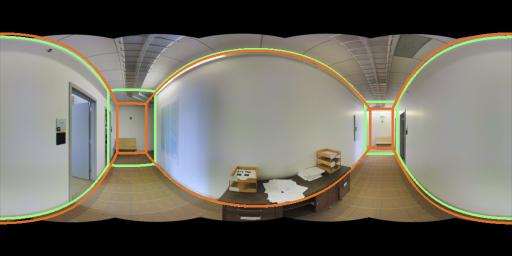}}&
\makecell{\includegraphics[width=0.23\linewidth]{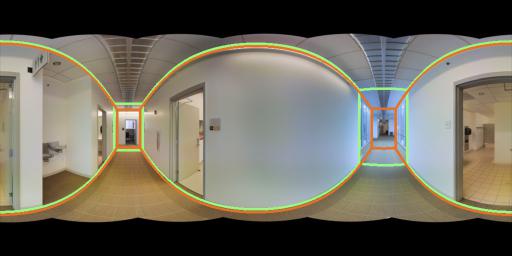}}&
\makecell{\includegraphics[width=0.23\linewidth]{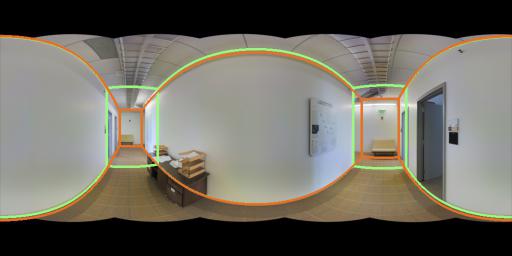}}&
\makecell{\includegraphics[width=0.23\linewidth]{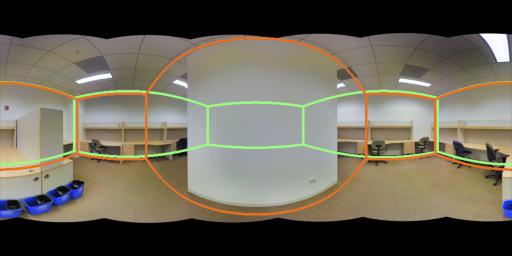}}\\
\end{tabular}
    \caption{
    Qualitative results of cuboid layout estimation on Stanford 2D-3D~\cite{2017arXiv170201105A} dataset.
    The results in the first to the fourth rows are separately sampled from four groups that comprise results with the best 0--25\%, 25--50\%, 50--75\% and 75-100\% corner errors, and the four results with the worst corner errors are displayed in the last row.
    The green lines are ground truth layout while the orange lines are estimated layout.
    }
\end{figure*}

\newpage
\section{More Qualitative Results of Non-Cuboid Room Layout Reconstruction}

\begin{figure*}[h]
   \centering
\setlength\tabcolsep{1.5pt}
\begin{tabular}{cc}
 \makecell{\includegraphics[height=3.2cm]{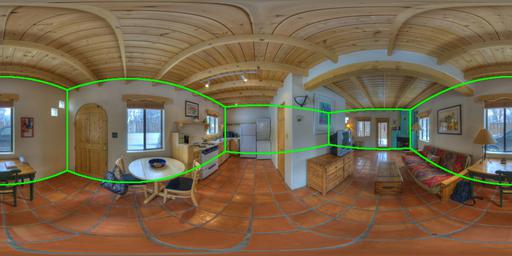} \\ \includegraphics[height=3.2cm]{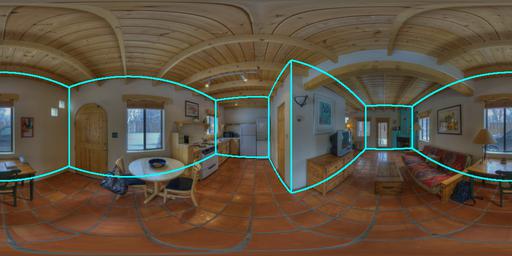}} & \makecell{\includegraphics[height=6.2cm]{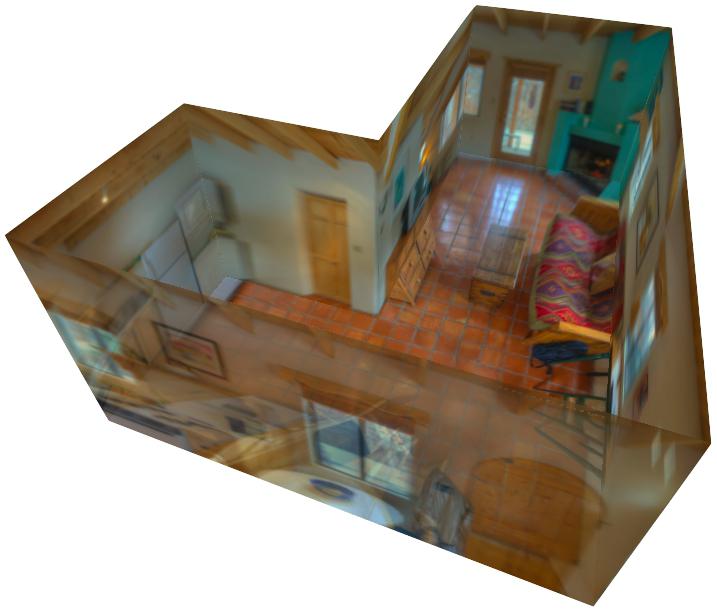}} \\
\end{tabular}
    \caption{Green lines are original ground truth annotation. Blue lines are room layout estimated by our model.}
\end{figure*}

\begin{figure*}[h]
   \centering
\setlength\tabcolsep{1.5pt}
\begin{tabular}{cc}
 \makecell{\includegraphics[height=3.2cm]{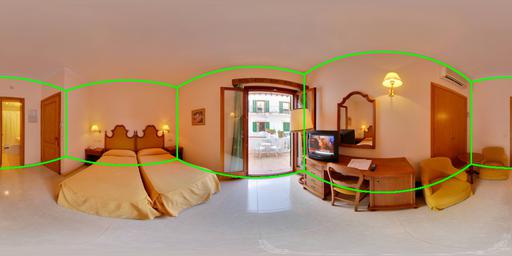} \\ \includegraphics[height=3.2cm]{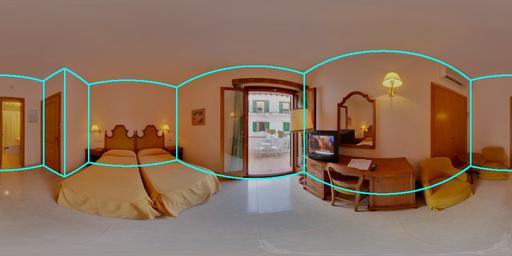}} & \makecell{\includegraphics[height=6.2cm]{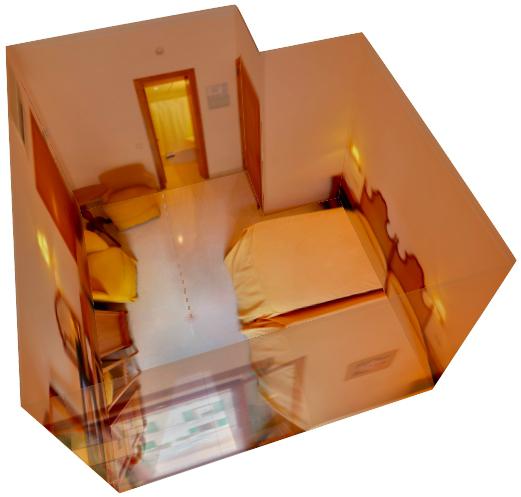}} \\
\end{tabular}
    \caption{Green lines are original ground truth annotation. Blue lines are room layout estimated by our model.}
\end{figure*}

\begin{figure*}[h]
   \centering
\setlength\tabcolsep{1.5pt}
\begin{tabular}{cc}
 \makecell{\includegraphics[height=3.2cm]{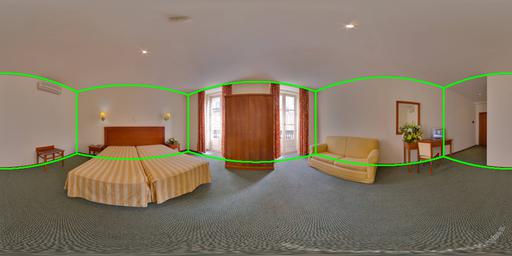} \\ \includegraphics[height=3.2cm]{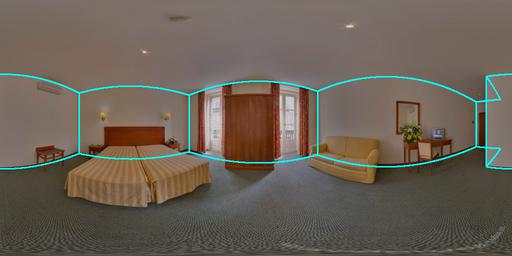}} & \makecell{\includegraphics[height=6.2cm]{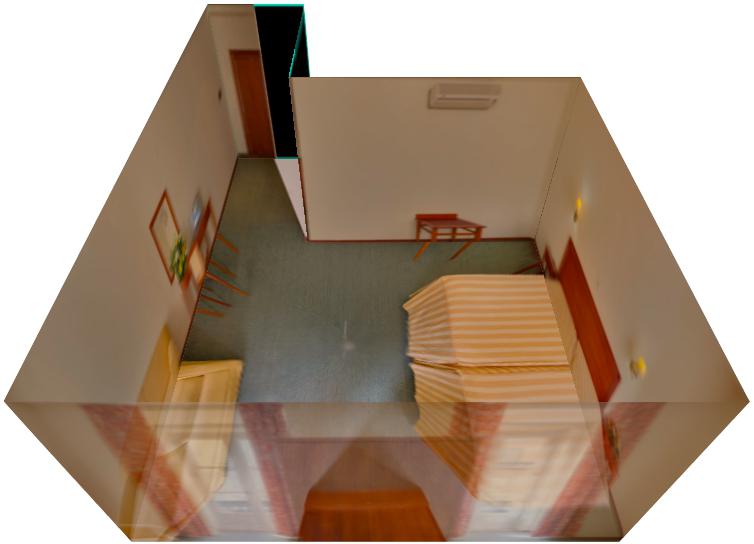}} \\
\end{tabular}
    \caption{Green lines are original ground truth annotation. Blue lines are room layout estimated by our model. The occlusion walls are filled with black.}
\end{figure*}

\begin{figure*}[h]
   \centering
\setlength\tabcolsep{1.5pt}
\begin{tabular}{cc}
 \makecell{\includegraphics[height=3.2cm]{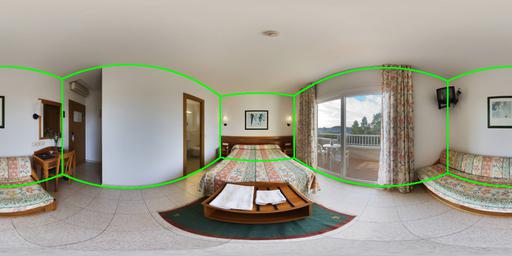} \\ \includegraphics[height=3.2cm]{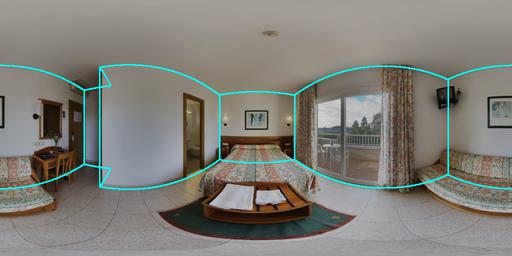}} & \makecell{\includegraphics[height=6.2cm]{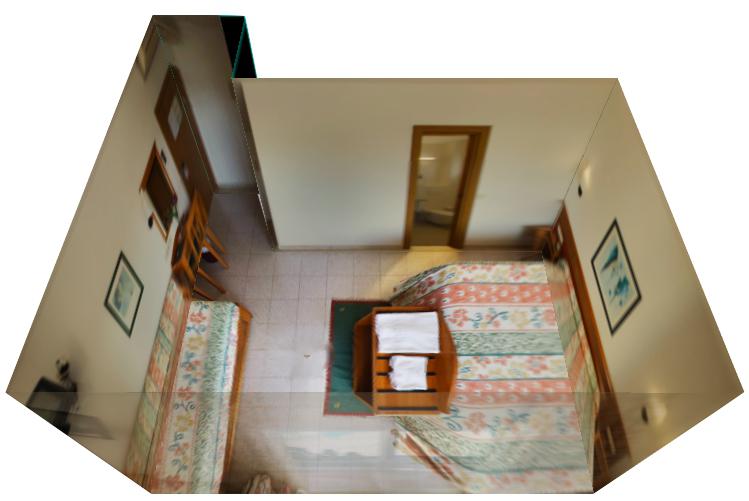}} \\
\end{tabular}
    \caption{Green lines are original ground truth annotation. Blue lines are room layout estimated by our model. The occlusion walls are filled with black.}
\end{figure*}

\begin{figure*}[h]
   \centering
\setlength\tabcolsep{1.5pt}
\begin{tabular}{cc}
 \makecell{\includegraphics[height=3.2cm]{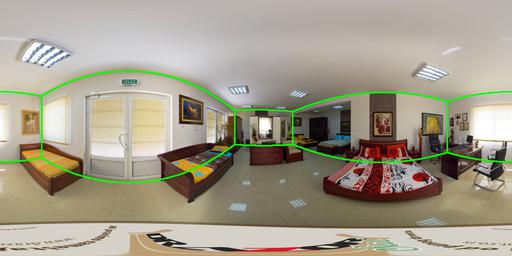} \\ \includegraphics[height=3.2cm]{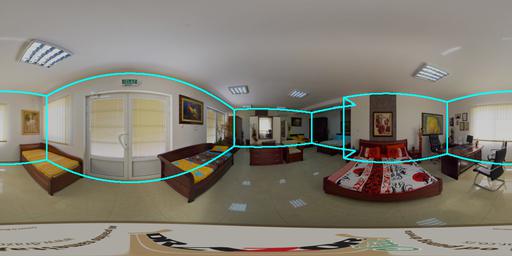}} & \makecell{\includegraphics[height=6.2cm]{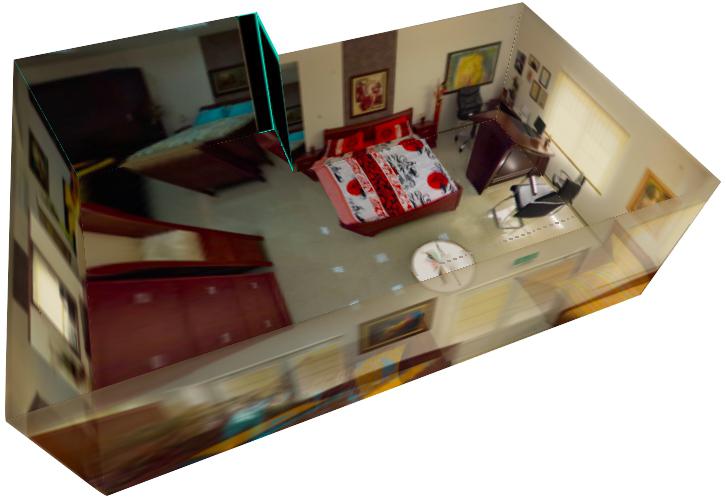}} \\
\end{tabular}
    \caption{Green lines are original ground truth annotation. Blue lines are room layout estimated by our model. The occlusion walls are filled with black.}
\end{figure*}

\begin{figure*}[h]
   \centering
\setlength\tabcolsep{1.5pt}
\begin{tabular}{cc}
 \makecell{\includegraphics[height=3.2cm]{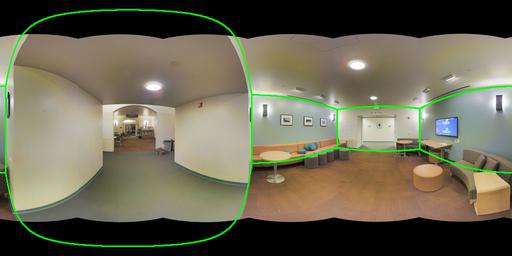} \\ \includegraphics[height=3.2cm]{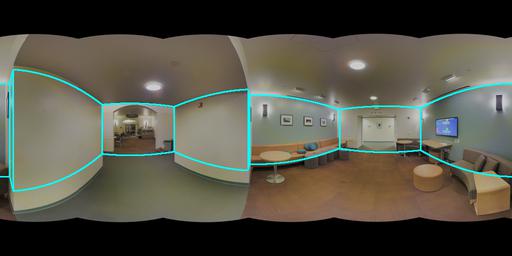}} & \makecell{\includegraphics[height=6.2cm]{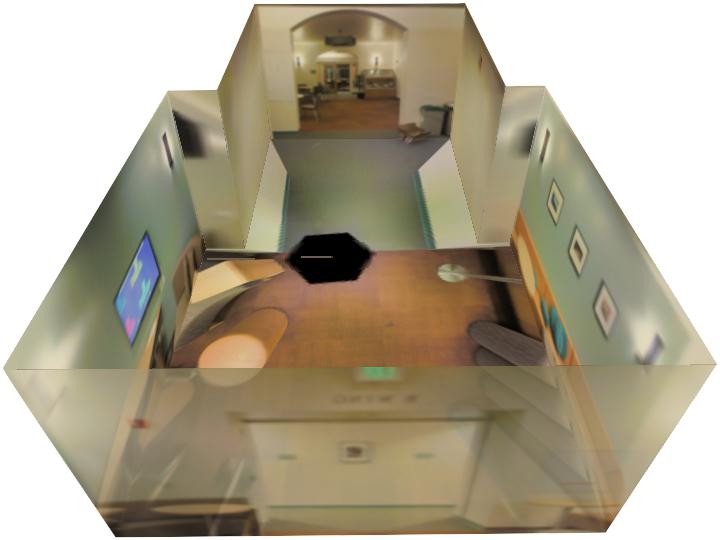}} \\
\end{tabular}
    \caption{Green lines are original ground truth annotation. Blue lines are room layout estimated by our model.}
\end{figure*}

\begin{figure*}[h]
   \centering
\setlength\tabcolsep{1.5pt}
\begin{tabular}{cc}
 \makecell{\includegraphics[height=3.2cm]{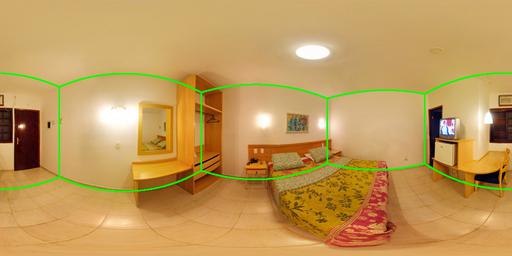} \\ \includegraphics[height=3.2cm]{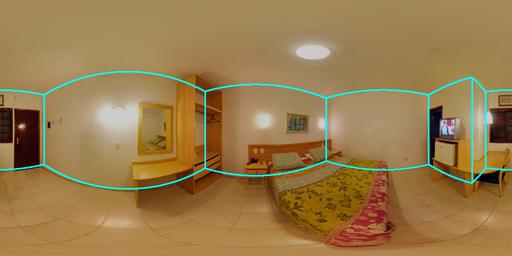}} & \makecell{\includegraphics[height=6.2cm]{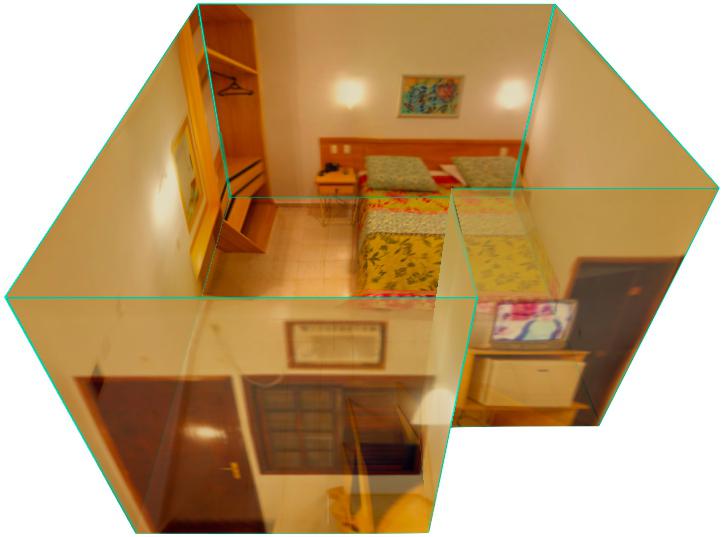}} \\
\end{tabular}
    \caption{Green lines are original ground truth annotation. Blue lines are room layout estimated by our model.}
\end{figure*}

\begin{figure*}[h]
   \centering
\setlength\tabcolsep{1.5pt}
\begin{tabular}{cc}
 \makecell{\includegraphics[height=3.4cm]{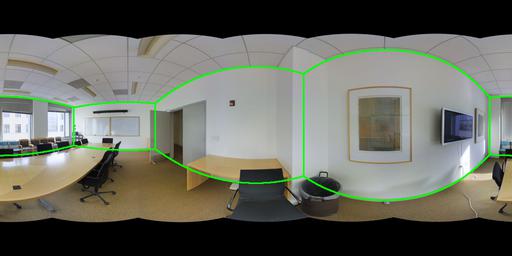} \\ \includegraphics[height=3.4cm]{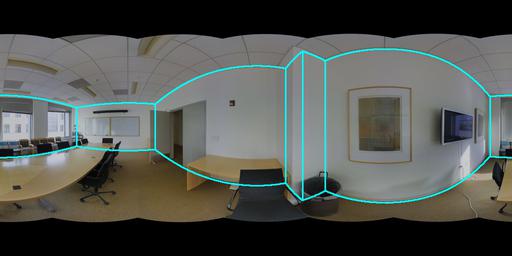}} & \makecell{\includegraphics[height=5.5cm]{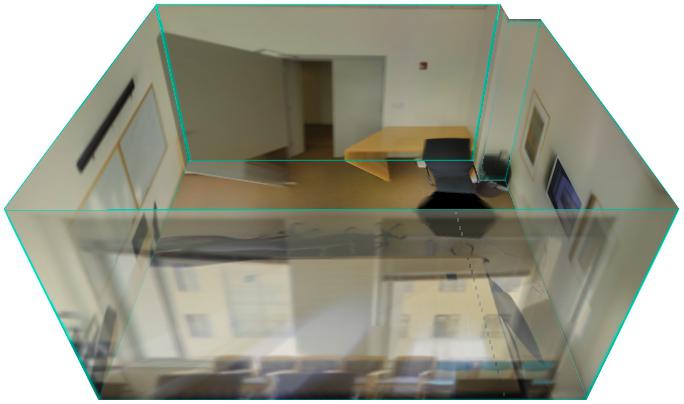}} \\
\end{tabular}
    \caption{Green lines are original ground truth annotation. Blue lines are room layout estimated by our model.}
\end{figure*}

\begin{figure*}[h]
   \centering
\setlength\tabcolsep{1.5pt}
\begin{tabular}{cc}
 \makecell{\includegraphics[height=3.4cm]{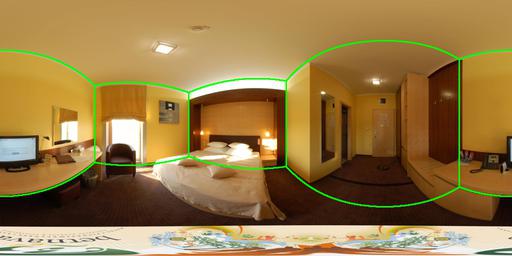} \\ \includegraphics[height=3.4cm]{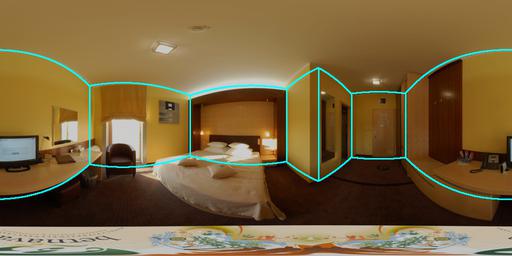}} & \makecell{\includegraphics[height=5.5cm]{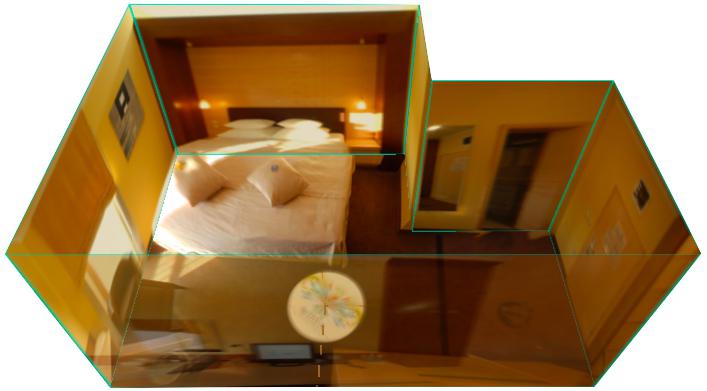}} \\
\end{tabular}
    \caption{Green lines are original ground truth annotation. Blue lines are room layout estimated by our model.}
\end{figure*}

\begin{figure*}[h]
   \centering
\setlength\tabcolsep{1.5pt}
\begin{tabular}{cc}
 \makecell{\includegraphics[height=3.4cm]{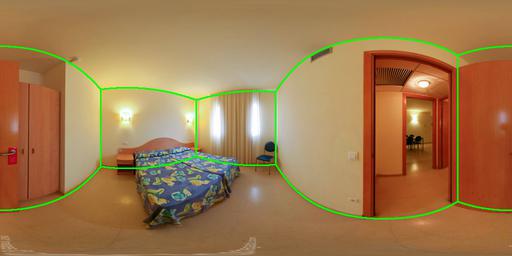} \\ \includegraphics[height=3.4cm]{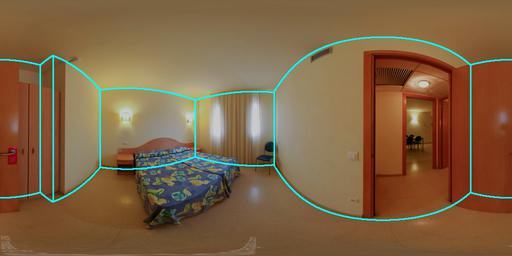}} & \makecell{\includegraphics[height=6.8cm]{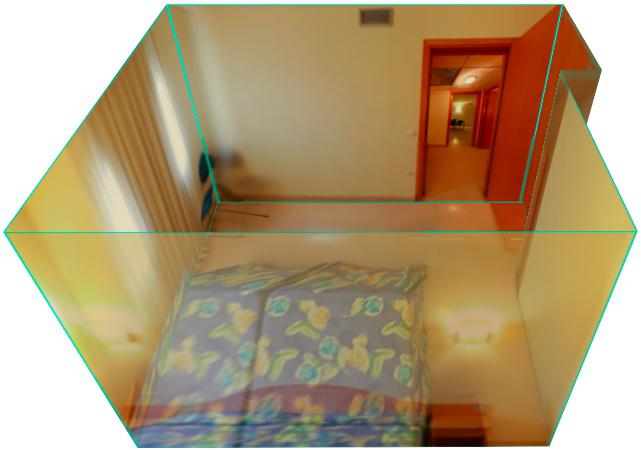}} \\
\end{tabular}
    \caption{Green lines are original ground truth annotation. Blue lines are room layout estimated by our model.}
\end{figure*}

\begin{figure*}[h]
   \centering
\setlength\tabcolsep{1.5pt}
\begin{tabular}{cc}
 \makecell{\includegraphics[height=3.4cm]{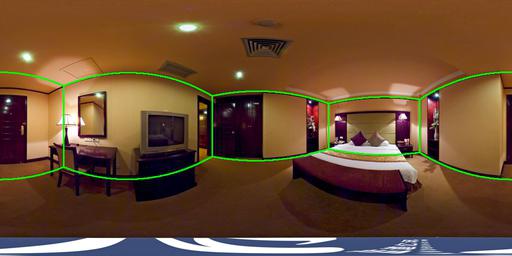} \\ \includegraphics[height=3.4cm]{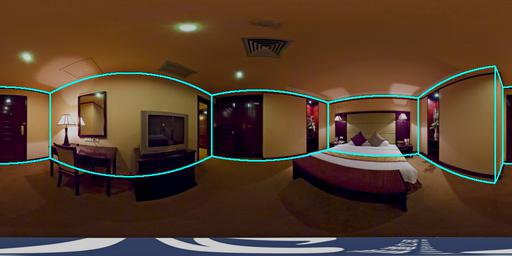}} & \makecell{\includegraphics[height=6.8cm]{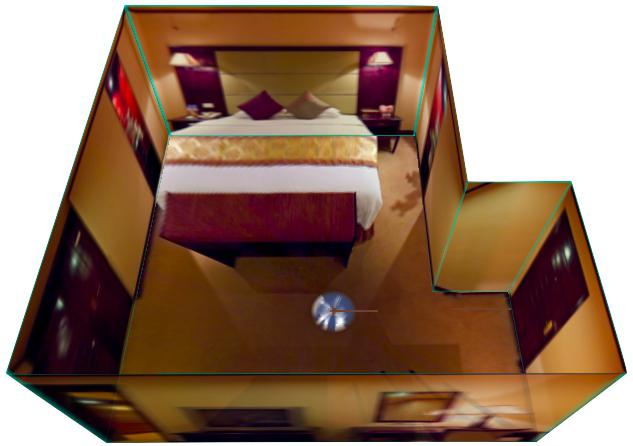}} \\
\end{tabular}
    \caption{Green lines are original ground truth annotation. Blue lines are room layout estimated by our model.}
\end{figure*}

\begin{figure*}[h]
   \centering
\setlength\tabcolsep{1.5pt}
\begin{tabular}{cc}
 \makecell{\includegraphics[height=3.4cm]{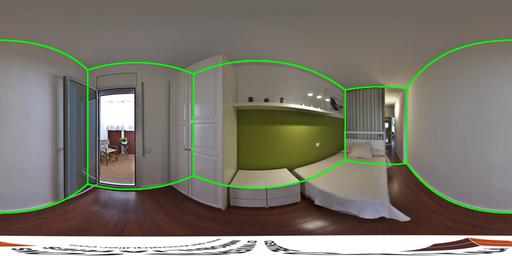} \\ \includegraphics[height=3.4cm]{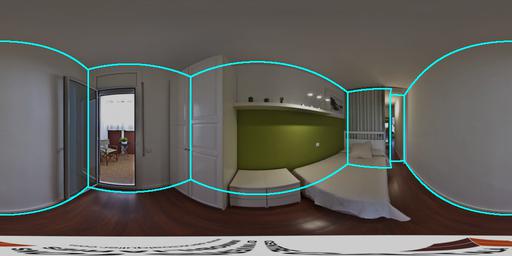}} & \makecell{\includegraphics[height=6.8cm]{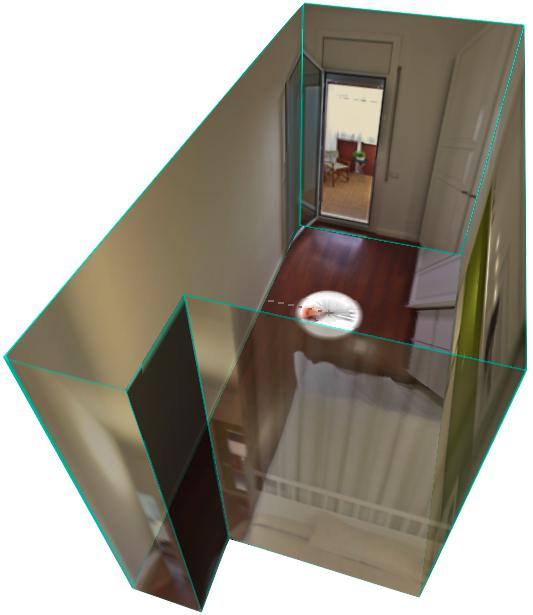}} \\
\end{tabular}
    \caption{Green lines are original ground truth annotation. Blue lines are room layout estimated by our model. The occlusion wall is filled with black.}
\end{figure*}

\begin{figure*}[h]
   \centering
\setlength\tabcolsep{1.5pt}
\begin{tabular}{cc}
 \makecell{\includegraphics[height=3.4cm]{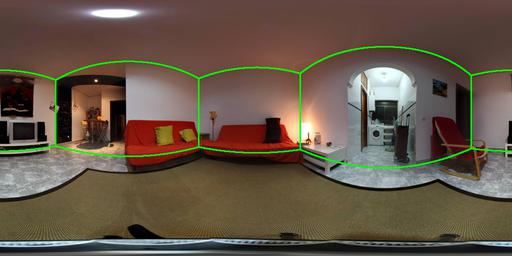} \\ \includegraphics[height=3.4cm]{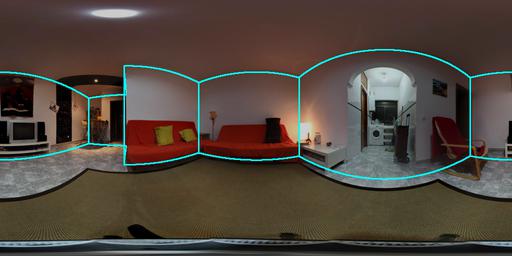}} & \makecell{\includegraphics[height=5.5cm]{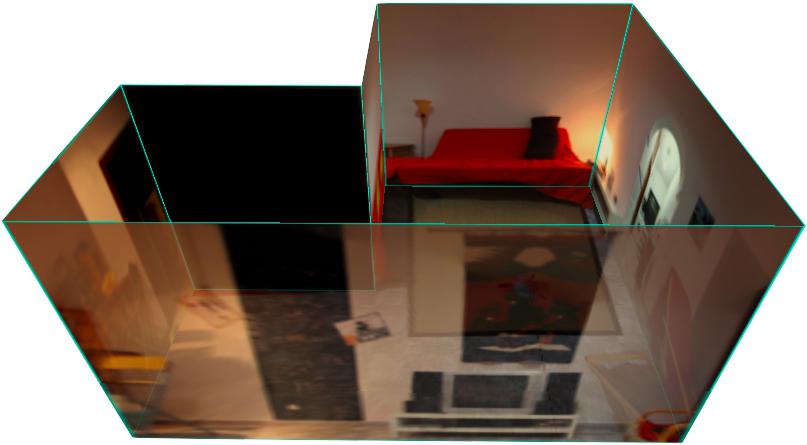}} \\
\end{tabular}
    \caption{Green lines are original ground truth annotation. Blue lines are room layout estimated by our model. The occlusion wall is filled with black.}
\end{figure*}

\end{document}